\documentclass[11pt,a4paper]{lumia}

\usepackage[numbers,sort&compress]{natbib}
\usepackage{graphicx}
\usepackage{xurl}
\usepackage{xspace}
\usepackage{array}
\usepackage{boldline}
\usepackage{float}
\usepackage{multirow}
\usepackage{makecell}
\usepackage{ragged2e}
\usepackage{mathtools}
\usepackage{subcaption}
\usepackage{wrapfig}
\usepackage{algorithm}
\usepackage{algpseudocode}
\algrenewcommand\algorithmicrequire{\textbf{Input:}}
\algrenewcommand\algorithmicensure{\textbf{Output:}}
\algrenewcommand\algorithmiccomment[1]{\hfill\mbox{\(\triangleright\) #1}}

\newcolumntype{C}[1]{>{\centering\arraybackslash}p{#1}}
\definecolor{gainred}{HTML}{C2601A}
\definecolor{cvprblue}{HTML}{C2601A}

\definecolor{medjevorange}{HTML}{C2601A}
\definecolor{medjevlink}{HTML}{A8500F}
\colorlet{lumiaabsbg}{medjevorange!8!white}
\renewcommand{\titlefont}{\color{medjevorange}\normalfont\bfseries\fontsize{16.5}{19}\selectfont}
\hypersetup{allcolors=medjevlink}

\makeatletter
\renewcommand{\maketitle}{%
\vspace*{-35pt}%
\bgroup\setlength{\parindent}{0pt}%
  \begin{adjustwidth}{0pt}{0pt}
    \begin{flushleft}
      {\centering \titlefont \@title\par}%
      \vskip9pt
      {\centering \@author\par}%
      \vskip6pt
    \end{flushleft}
  \end{adjustwidth}
\egroup
\vskip-8pt%
\noindent\textcolor{black}{\rule{\linewidth}{1pt}}%
\vskip4pt
{\abscontent}%
\thispagestyle{firststyle}%
}
\makeatother

\setheadertext{\textsc{MedJev}}
\setheadertitle{\textsc{MedJev}}
\setreportdate{\raisebox{0pt}{\includegraphics[height=1.75em]{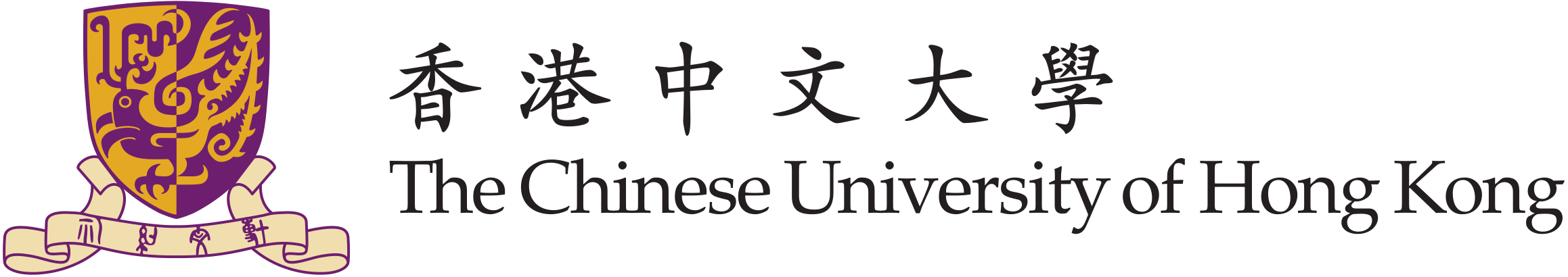}}}

\title{\textsc{MedJev}: A System One Vision-Language Decision Model\\ for Fast and Effective Medical Image Understanding}

\author{
{\fontsize{10}{12}\selectfont
\mbox{Fan Zhang\textsuperscript{1}},\hspace{0.3em}
\mbox{Hao Wu\textsuperscript{2}},\hspace{0.3em}
\mbox{Jinyang Wu\textsuperscript{2}},\hspace{0.3em}
\mbox{Hongyu Qu\textsuperscript{3}}
\par
\vspace{-6pt}

\mbox{Zheng Lian\textsuperscript{4}},\hspace{0.3em}
\mbox{Dongfu Yin\textsuperscript{5}},\hspace{0.3em}
\mbox{Qingsong Wen\textsuperscript{6}},\hspace{0.3em}
\mbox{Pheng-Ann Heng\textsuperscript{1}}
}
\par
\vspace{3pt}

{\normalfont\fontsize{8}{10}\selectfont
\textsuperscript{1}The Chinese University of Hong Kong
\quad
\textsuperscript{2}Tsinghua University
\quad
\textsuperscript{3}National University of Singapore
\quad
\textsuperscript{4}Tongji University
\par
\textsuperscript{5}Guangdong Laboratory of Artificial Intelligence and Digital Economy (SZ)
\quad
\textsuperscript{6}Squirrel Ai Learning
}
\par\vspace{-6pt}
}

\begin{document}

\begin{abstract}
Medical image understanding with vision-language models (VLMs) typically relies on autoregressively generating reasoning and answer tokens.
These unnecessary tokens make the inference cost scale with the response length and force the answer to be parsed from free-form text, incurring avoidable latency and decision errors.
To this end, we introduce \textsc{MedJev}, a \emph{system one} vision-language decision model that maps medical images, questions, and candidate answers directly to decisions in a single multimodal forward pass, without any autoregressive generation process.
For fast inference, \textsc{MedJev} removes the upper language layers and the generative output pathway, while keeping the full vision pathway and all visual tokens.
On this shortened backbone, a decision head scores each candidate by pretrained query--key matching and refines the scores with a lightweight multilayer perceptron (MLP) residual, coupling the pretrained matching structure with task-specific nonlinear discrimination.
Built on Qwen3.5-4B, \textsc{MedJev}-4B improves both accuracy and speed over the original model on each of four medical image classification and visual question answering benchmarks, with an absolute accuracy gain of \textbf{7.93\%} and a \textbf{6.04$\times$} end-to-end speedup.
Further comparison with 20 cutting-edge models shows that \textsc{MedJev}-4B achieves state-of-the-art performance and inference speed among 4B--9B general and medical VLMs, indicating that direct multimodal decision making is an effective route to fast and effective medical image understanding.

\par\medskip
\noindent{\faGithub}\, \textbf{Code:}~\url{https://github.com/zfkarl/MedJev}
\end{abstract}

\maketitle

\section{Introduction}
\label{sec:introduction}

\begin{figure}[htbp]
\includegraphics[height=2.05in]{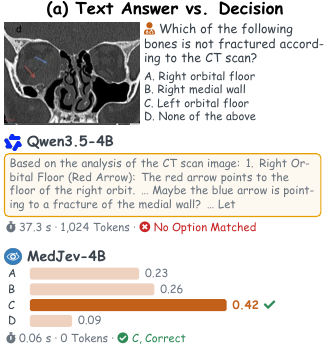}\hfill
\raisebox{0.08in}{\textcolor[HTML]{C8CDD2}{\rule{0.5pt}{1.90in}}}\hfill
\includegraphics[height=2.05in]{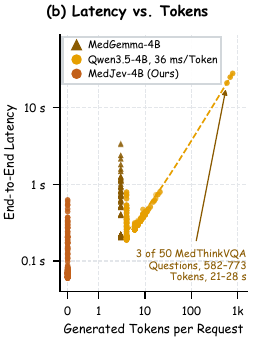}\hfill
\raisebox{0.08in}{\textcolor[HTML]{C8CDD2}{\rule{0.5pt}{1.90in}}}\hfill
\includegraphics[height=2.05in]{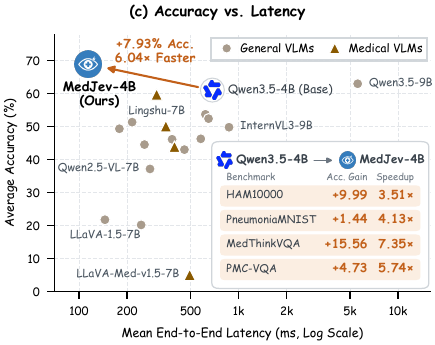}
\caption{(a) An example from PMC-VQA~\cite{zhang2023pmcvqa}, where Qwen3.5-4B writes 1,024 tokens without selecting an option and \textsc{MedJev}-4B scores the candidates in one forward pass. (b) Per-request latency against generated tokens, thinking off. (c) Average accuracy and latency on four benchmarks, where \textsc{MedJev}-4B beats all 20 general and medical VLMs at the 4B--9B scale on both.}
\label{fig:teaser}
\end{figure}

Vision-language models (VLMs) have become a general interface for understanding medical images.
Both general-purpose VLMs~\cite{liu2023llava,zhu2025internvl3,wang2025internvl35,bai2025qwen25vl,bai2025qwen3vl} and medical VLMs trained on biomedical image-text data~\cite{li2023llavamed,chen2024huatuogptvision,xu2025lingshu,sellergren2025medgemma} can read dermoscopy images, radiographs, and biomedical figures and answer clinical questions about them in natural language.
Going one step further, recent medical VLMs are trained to write out their reasoning traces before giving the final answer~\cite{pan2025medvlmr1,lai2025medr1}.
For all of these models, obtaining the result of a medical image task relies on autoregressive generation and decoding.

This autoregressive process is an expensive and indirect way to obtain these results.
Each generated token requires another forward pass through the language model, so the inference cost grows with the length of the response (Fig.~\ref{fig:teaser} (b)), and reasoning traces make the response much longer.
Even with thinking disabled, the original Qwen3.5-4B~\cite{qwen2026qwen35} answers a typical question of the multi-image benchmark MedThinkVQA~\cite{yao2026medicalthinking} in 0.35\,s in our measurements, yet its 95th-percentile latency exceeds 20\,s and its mean latency reaches 1.84\,s.
Beyond this cost, the decision has to be parsed from free-form text with rules or an extra model, and a response that matches none of the candidates is counted as an error (Fig.~\ref{fig:teaser} (a)).
Moreover, the reasoning traces contribute little, as more than 70\% of model errors on MedThinkVQA come from reading the images and integrating evidence across views, and and longer traces help little once this grounding is weak~\cite{yao2026medicalthinking}.

The above two problems both stem from producing the decision as generated text.
Dual-process theories of cognition describe two modes of thinking, a fast and intuitive \emph{system one} and a slow and deliberate \emph{system two}~\cite{kahneman2011thinking,evans2013dualprocess}.
Inspired by this theory, Jev models~\cite{almeida2026jev} have quickly become popular, mainly because they give up free-form generation and return typed and probabilistic decisions directly, which removes both the decoding cost and the output parsing.
Many medical image tasks ask for one choice among a fixed set of diagnoses or answers, which matches this form of decision making.
We therefore aim to understand medical images in a system one manner and propose \textsc{MedJev}-4B, a fast and effective vision-language decision model built on Qwen3.5-4B.
Given one or more medical images, a question, and a set of candidate answers, \textsc{MedJev} appends a marker token to each candidate and a decision token to the end of the input, scores all candidates jointly in a single forward pass, and takes the highest-scoring candidate as its prediction.
Since no reasoning or answer tokens are generated, the decision cost no longer grows with the response length, and a prediction is always valid.

The core idea of \textsc{MedJev} is to make the decision with less language-side computation while keeping all visual evidence and the matching structure learned in pretraining.
It realizes this idea with two complementary designs, a language-side depth reduction that shortens the path from the visual evidence to the decision for speed, and a residual decision head that refines pretrained query--key matching for accuracy on the heterogeneous medical decisions.

\textbf{Language-Side Depth Reduction.}
Starting from a pretrained VLM, \textsc{MedJev} keeps the vision encoder, the vision-language connector, and every visual token that they produce.
Since studies of large language models find that many of the deepest layers can be removed with limited loss in performance~\cite{men2024shortgpt,gromov2025unreasonable}, it removes the language-side computation that the decision does not use, namely the upper layers before the last one, the value, gate, output, and feed-forward paths of the last layer, and the vocabulary projection.
In \textsc{MedJev}-4B, these components account for 20\% of the parameters.
Visual token pruning methods~\cite{chen2024fastv,zhang2025sparsevlm,xing2025pyramiddrop} accelerate VLMs by discarding encoded visual tokens, which may remove small lesions and fine structures.
\textsc{MedJev} therefore keeps the visual evidence intact and shortens only the language-side path to the decision.

\textbf{Residual Decision Head.}
To score the candidates, \textsc{MedJev} reuses the query and key projections of the last layer, which already learned in pretraining how strongly one token attends to another.
It maps the decision token to a query and each candidate marker to a key, and takes their scaled dot product as the matching score of the candidate.
Because pretrained matching alone may not be expressive enough for heterogeneous medical images and questions, a lightweight multilayer perceptron (MLP) shared across candidates adds a residual score computed from the decision state, the candidate state, and their elementwise product.
Its output layer is initialized to zero, so training starts from pretrained matching and learns task-specific nonlinear discrimination with few new parameters.

Extensive experiments on four medical image understanding benchmarks that cover dermoscopy, chest radiographs, multi-image clinical cases, and biomedical figures~\cite{tschandl2018ham10000,yang2023medmnist,yao2026medicalthinking,zhang2023pmcvqa} demonstrate the superiority of \textsc{MedJev}-4B over the competing baselines.
Compared with the original Qwen3.5-4B, \textsc{MedJev}-4B improves the average accuracy by an absolute 7.93\%, from 61.13\% to 69.06\%, and accelerates end-to-end inference by 6.04$\times$, from 685.03\,ms to 113.44\,ms, with accuracy gains on every benchmark.
These gains need no task-specific output head, because a single model scores candidate sets that range from a binary chest radiograph decision to a seven-way dermoscopic diagnosis.
The speedup tracks what generation would have cost, so it peaks on the multi-image benchmark, whose cases drive a generative model to the longest responses of the four.
In addition, comparison with 20 off-the-shelf general and medical VLMs (Fig.~\ref{fig:teaser} (c)) shows that \textsc{MedJev}-4B is the only model at the 4B--9B scale that is both the most accurate and the fastest on average.

The main contribution of this paper is threefold:
\begin{itemize}
    \item We introduce \textsc{MedJev}, a system one vision-language decision model that maps medical images, questions, and candidate answers directly to decisions in a single forward pass, without any autoregressive generation.
    \item We design a language-side depth reduction that removes the computation a decision does not need while keeping every visual token, and a residual decision head that refines pretrained query--key matching with a lightweight zero-initialized MLP for accurate decisions.
    \item On four medical image understanding benchmarks, \textsc{MedJev}-4B improves the average accuracy of Qwen3.5-4B by an absolute 7.93\% with a 6.04$\times$ end-to-end speedup, and achieves state-of-the-art accuracy and speed among 4B--9B general-purpose and medical VLMs.
\end{itemize}

\section{Related Work}
\label{sec:related}

\subsection{Medical Vision-Language Models}
Medical VLMs largely inherit the generative design of general-purpose VLMs and adapt it to clinical data.
Early models tune LLaVA~\cite{liu2023llava} on biomedical figure-caption pairs~\cite{li2023llavamed}, learn from interleaved medical images and text~\cite{moor2023medflamingo}, or train one generalist model across many medical tasks and modalities~\cite{tu2023medpalmm,wu2023radfm}.
Later models scale up instruction data and backbones~\cite{chen2024huatuogptvision,xu2025lingshu,sellergren2025medgemma}, while strong general-purpose VLMs~\cite{bai2025qwen25vl,bai2025qwen3vl,wang2025internvl35,li2024llavaonevision} have also become competitive baselines on medical tasks.
Inspired by chain-of-thought prompting~\cite{wei2022cot} and reinforcement learning with verifiable rewards~\cite{deepseek2025r1}, recent work further trains medical VLMs to write out reasoning steps before the answer~\cite{pan2025medvlmr1,lai2025medr1}.
These models are evaluated on benchmarks that range from a mix of closed and open questions~\cite{lau2018vqarad,liu2021slake,he2020pathvqa} to multiple-choice questions over given candidates~\cite{zhang2023pmcvqa,hu2024omnimedvqa,yao2026medicalthinking}, yet all of them deliver their decisions as generated text.
Because the decision arrives inside a sentence, these benchmarks recover it by matching strings or by asking a separate judge model, and a response that states the right finding in an unexpected form can still be scored wrong.
Contrastive models such as BiomedCLIP~\cite{zhang2023biomedclip} instead score an image against text labels directly, but they encode the two modalities separately and cannot condition the scores on a question or on several images.
\textsc{MedJev} combines both families by keeping the joint multimodal processing of a VLM while replacing text generation with a direct decision.

\subsection{Efficient Vision-Language Inference}
The inference cost of a VLM comes mainly from long visual token sequences, the depth of the language model, and autoregressive decoding.
Token reduction methods discard or merge visual tokens~\cite{chen2024fastv,zhang2025sparsevlm,xing2025pyramiddrop,shang2025prumerge,yang2024visionzip}, which shortens the input but risks removing small medical findings.
Layer pruning methods remove redundant deep layers of large language models~\cite{men2024shortgpt,gromov2025unreasonable}, yet the pruned models still generate their answers token by token.
Decoding also leaves the cost of a single answer open, since a model may spend a few tokens or several hundred on the same question.
A generated answer carries the decision inside open-ended text, so the decision shares the parameters and the computation of captioning and dialogue.
System one models~\cite{almeida2026jev} instead remove decoding and return typed decisions, and concurrent work extends them to images~\cite{yu2026visualjev,zhou2026pixeljev}, videos~\cite{zhu2026videojev}, spatial reasoning~\cite{qin2026mjevspatial}, histopathology~\cite{tommasino2026pathjev}, and ophthalmic cases~\cite{liu2026retijev}.
Among them, MJev-Spatial~\cite{qin2026mjevspatial} reads candidate scores from the native query and key projections at the top of the full language stack, while earlier work scores option letters with the language model head~\cite{robinson2023mcqa}.
\textsc{MedJev} removes decoding and the upper language layers together, keeps every visual token, and and refines the pretrained query--key matching with a zero-initialized residual head.

\begin{figure}[t]
\centering
\includegraphics[width=\textwidth]{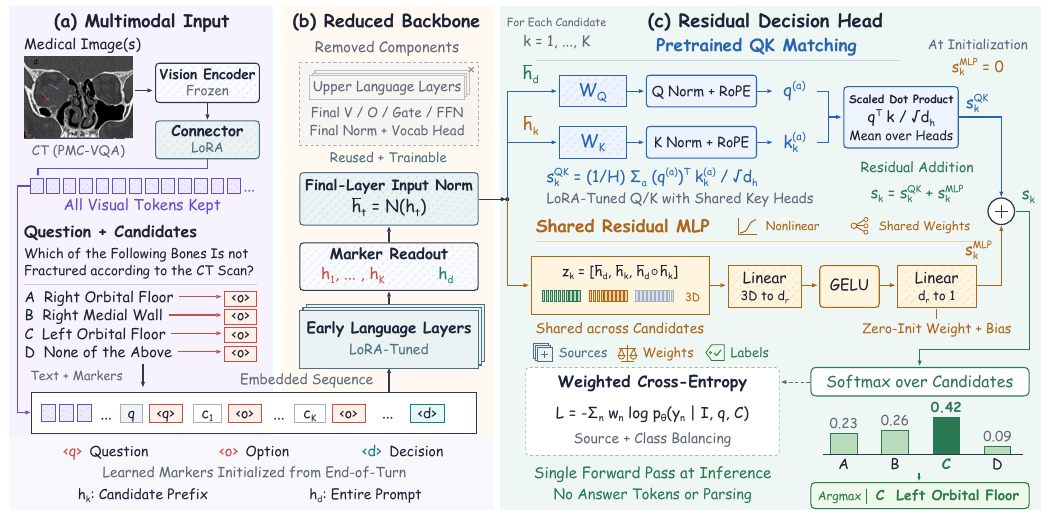}
\caption{\textbf{Overview of \textsc{MedJev}.} (a) The images, the question, and all candidates form a single prompt with learned markers, and every visual token is kept. (b) A reduced backbone returns the marker states. (c) Pretrained query--key matching and a shared MLP residual score the candidates in one pass. The example is a CT question of PMC-VQA~\cite{zhang2023pmcvqa} with the probabilities of \textsc{MedJev}-4B.}
\label{fig:overview}
\end{figure}

\section{Methodology}
\label{sec:method}

\textsc{MedJev} turns a pretrained VLM into a decision model that reads the images, the question, and all candidate answers in one pass and returns a score for each candidate (Fig.~\ref{fig:overview}).
It formulates medical image understanding as a selection among the given candidates (Sec.~\ref{sec:formulation}), shortens the language side of the backbone to the computation that this selection needs (Sec.~\ref{sec:depth}), and scores the candidates with a residual decision head built on pretrained query--key matching (Sec.~\ref{sec:head}).
The whole model is trained with a single cross-entropy objective and runs with no decoding loop and no parser (Sec.~\ref{sec:training}), and Algorithm~\ref{alg:medjev} summarizes the full training and inference procedure.

\subsection{Decision Formulation}
\label{sec:formulation}
Each example contains $M\geq 1$ medical images $\mathcal{I}=\{I_1,\dots,I_M\}$, a question $q$, and $K$ candidate answers $\mathcal{C}=\{c_1,\dots,c_K\}$, and the goal is to predict the index $y\in\{1,\dots,K\}$ of the correct candidate.
The candidates are written in text and can be diagnostic labels, findings, or answer options, so classification and multiple-choice VQA share the same form.
A generative VLM solves this task indirectly.
It produces a response $\mathbf{o}=(o_1,\dots,o_T)$ token by token:
\begin{equation}
p(\mathbf{o}\mid\mathcal{I},q,\mathcal{C})=\prod_{t=1}^{T}p(o_t\mid o_{<t},\mathcal{I},q,\mathcal{C}),
\label{eq:ar}
\end{equation}
and a parser then maps $\mathbf{o}$ to a candidate, so the cost grows with $T$ and the prediction fails whenever $\mathbf{o}$ names no candidate.
\textsc{MedJev} instead computes a score vector $\mathbf{s}=f_\theta(\mathcal{I},q,\mathcal{C})\in\mathbb{R}^{K}$ in a single forward pass and predicts:
\begin{equation}
\hat{y}=\operatorname*{arg\,max}_{k\in\{1,\dots,K\}} s_k ,
\label{eq:argmax}
\end{equation}
which is always a given candidate.

To compute $\mathbf{s}$ in one pass, \textsc{MedJev} writes the question and all candidates into a single prompt in the chat template of the backbone and adds three special tokens to its vocabulary.
A question marker $\langle q\rangle$ follows the question, an option marker $\langle o\rangle$ follows each candidate, and a decision marker $\langle d\rangle$ closes the prompt after an instruction to select the best answer, so that the language model receives:
\begin{equation}
\big[\,V_1,\dots,V_M,\;q,\;\langle q\rangle,\;c_1,\;\langle o\rangle,\;\dots,\;c_K,\;\langle o\rangle,\;\langle d\rangle\,\big],
\label{eq:input}
\end{equation}
where $V_m$ denotes the visual tokens of image $I_m$ and each candidate is preceded by its option letter.
The marker embeddings are initialized from the end-of-turn token of the backbone and learned during training.
Let $t_k$ be the position of the option marker after $c_k$ and $t_d$ the position of the decision marker.
Under causal attention, the hidden state at $t_k$ summarizes the images, the question, and the candidates up to $c_k$, while the hidden state at $t_d$ summarizes the entire input.
These marker states are the only outputs that the decision requires from the language model.

\needspace{20\baselineskip}
\subsection{Language-Side Depth Reduction}
\label{sec:depth}
\begin{wrapfigure}{r}{0.47\textwidth}
\vspace{-\intextsep}
\centering
\includegraphics[width=0.45\textwidth]{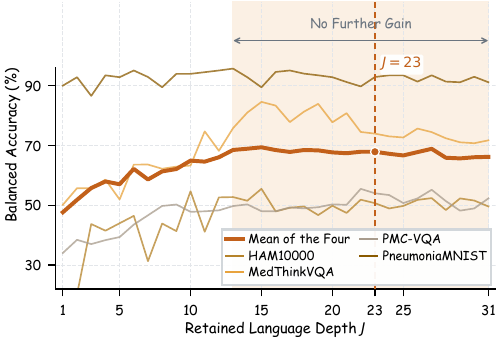}
\caption{\textbf{Depth probe analysis.} Balanced accuracy of the pretrained query-key match combined with a linear head over the marker states of each pretrained layer, with the backbone frozen and all examples from the training split.}
\label{fig:depth_probe}
\vspace{-\intextsep}
\end{wrapfigure}
A generative VLM runs its full language stack and the vocabulary projection at every decoding step, whereas a decision only needs the hidden states at the marker positions.
\textsc{MedJev} exploits this asymmetry.
A probe of the frozen backbone confirms this, since the marker states stop becoming more discriminative well before the last layer (Fig.~\ref{fig:depth_probe}).
Consider a backbone whose vision encoder and vision-language connector map each image to visual tokens, and whose language model stacks $L$ layers $F_1,\dots,F_L$ with an attention layer at the top.
\textsc{MedJev} keeps the vision encoder, the connector, and the first $J$ language layers unchanged and removes layers $J{+}1$ to $L{-}1$, so the states that reach the decision head are:
\begin{equation}
\mathbf{h}_t=\big[(F_J\circ\cdots\circ F_1)(\mathbf{X})\big]_t,\quad t\in\{t_1,\dots,t_K,t_d\},
\label{eq:depth}
\end{equation}
where $\mathbf{X}$ is the embedded input sequence of Eq.~\eqref{eq:input} and $[\cdot]_t$ takes the $D$-dimensional state at position $t$.
From the last layer $F_L$, \textsc{MedJev} keeps only the input normalization, the query and key projections with their normalizations, and the rotary position embedding~\cite{su2021roformer}, which the decision head reuses.
The value and output projections, any attention gate, and the feed-forward network of $F_L$ are removed together with the final normalization and the vocabulary projection, so the language model executes $J{+}1$ modules instead of $L$.
Every visual token is kept, and none is pruned or merged inside the language model, so the reduction shortens only the language-side path to the decision.

\subsection{Residual Decision Head}
\label{sec:head}
The last layer was pretrained to measure how strongly one token attends to another, which matches the relation that a decision needs between its marker and each candidate.
\textsc{MedJev} therefore scores the candidates with the retained query and key projections.
Let $\mathrm{N}(\cdot)$ be the input normalization of the last layer, and let $\bar{\mathbf{h}}_d=\mathrm{N}(\mathbf{h}_{t_d})$ and $\bar{\mathbf{h}}_k=\mathrm{N}(\mathbf{h}_{t_k})$.
For each query head $a\in\{1,\dots,H\}$, the head computes:
\begin{equation}
\mathbf{q}^{(a)}=\mathrm{R}_{t_d}\mathrm{N}_{Q}(\mathbf{W}_{Q}^{(a)}\bar{\mathbf{h}}_d),\;
\mathbf{k}_k^{(a)}=\mathrm{R}_{t_k}\mathrm{N}_{K}(\mathbf{W}_{K}^{(g)}\bar{\mathbf{h}}_k),
\label{eq:qk}
\end{equation}
where $\mathbf{W}_Q^{(a)}$ and $\mathbf{W}_K^{(g)}$ are the pretrained query and key projections, $g=g(a)$ is the key head shared by query head $a$ under grouped-query attention~\cite{ainslie2023gqa}, $\mathrm{N}_Q$ and $\mathrm{N}_K$ are the query and key normalizations, and $\mathrm{R}_t$ applies the rotary embedding at position $t$.
The matching score of candidate $k$ averages the scaled dot products over heads:
\begin{equation}
s_k^{\mathrm{QK}}=\frac{1}{H}\sum_{a=1}^{H}\frac{\big\langle\mathbf{q}^{(a)},\mathbf{k}_k^{(a)}\big\rangle}{\sqrt{d_h}},
\label{eq:qkscore}
\end{equation}
where $d_h$ is the head dimension.
This score is the attention logit of the last layer and is free.

Pretrained matching alone is not sufficient in this setting.
The query and key projections were trained on the output of the penultimate layer, yet after depth reduction they receive the output of layer $J$, and a bilinear score may not capture every interaction that a medical decision requires.
\textsc{MedJev} therefore adds a nonlinear residual computed by a lightweight MLP shared across candidates:
\begin{equation}
s_k^{\mathrm{MLP}}=\mathbf{w}_2^{\top}\,\mathrm{GELU}\big(\mathbf{W}_1\mathbf{z}_k+\mathbf{b}_1\big)+b_2,
\label{eq:mlp}
\end{equation}
where $\mathbf{z}_k\in\mathbb{R}^{3D}$ concatenates $\bar{\mathbf{h}}_d$, $\bar{\mathbf{h}}_k$, and their elementwise product $\bar{\mathbf{h}}_d\odot\bar{\mathbf{h}}_k$, and $\mathbf{W}_1$ projects it to a small hidden dimension.
The final score of candidate $k$ is:
\begin{equation}
s_k=s_k^{\mathrm{QK}}+s_k^{\mathrm{MLP}}.
\label{eq:score}
\end{equation}
We initialize $\mathbf{w}_2$ and $b_2$ to zero, so that $s_k=s_k^{\mathrm{QK}}$ at the start of training and the MLP learns only the correction that pretrained matching misses.
Because all candidates are scored with the same parameters, the decision head is not tied to a fixed label set and handles any number of candidates.

\subsection{Training and Inference}
\label{sec:training}
\begin{algorithm}[t]
\caption{Training and Inference of \textsc{MedJev}}
\label{alg:medjev}
\small
\begin{algorithmic}[1]
\Require Pretrained VLM with language layers $F_1,\dots,F_L$, depth $J$, training set $\mathcal{D}=\{(\mathcal{I}_n,q_n,\mathcal{C}_n,y_n)\}$ with weights $w_n$
\Ensure Prediction $\hat{y}$ for a test input $(\mathcal{I},q,\mathcal{C})$
\State Add the markers $\langle q\rangle,\langle o\rangle,\langle d\rangle$ to the vocabulary \Comment{Sec.~\ref{sec:formulation}}
\State Remove $F_{J+1},\dots,F_{L-1}$ \Comment{Sec.~\ref{sec:depth}}
\State Reduce $F_L$ to its query and key branches
\State Add a residual MLP with zero-initialized output \Comment{Sec.~\ref{sec:head}}
\Function{$f_\theta$}{$\mathcal{I},q,\mathcal{C}$}
  \State $\mathbf{X}\gets$ embedded input sequence of Eq.~\eqref{eq:input}
  \State $\mathbf{h}_t\gets\big[(F_J\circ\cdots\circ F_1)(\mathbf{X})\big]_t$ for $t\in\{t_1,\dots,t_K,t_d\}$
  \State \Return $\mathbf{s}$ with $s_k=s_k^{\mathrm{QK}}+s_k^{\mathrm{MLP}}$ \Comment{Eqs.~\eqref{eq:qk} to \eqref{eq:score}}
\EndFunction
\For{each minibatch $\mathcal{B}\subset\mathcal{D}$} \Comment{Training}
  \State $\mathbf{s}_n\gets f_\theta(\mathcal{I}_n,q_n,\mathcal{C}_n)$ for $n\in\mathcal{B}$
  \State $\mathcal{L}_{\mathcal{B}}\gets-\sum_{n\in\mathcal{B}}w_n\log\mathrm{softmax}(\mathbf{s}_n)_{y_n}$ \Comment{Eq.~\eqref{eq:loss}}
  \State Update LoRA, markers, norms of $F_L$, and MLP with $\nabla\mathcal{L}_{\mathcal{B}}$
\EndFor
\State Merge LoRA and markers into the weights \Comment{Deployment}
\State \Return $\hat{y}=\arg\max_k f_\theta(\mathcal{I},q,\mathcal{C})_k$ \Comment{Inference}
\end{algorithmic}
\end{algorithm}
A single \textsc{MedJev} model is trained on a mixture of medical classification and VQA sources (Algorithm~\ref{alg:medjev}).
For each example, the final scores are normalized with a softmax:
\begin{equation}
p_\theta(k\mid\mathcal{I},q,\mathcal{C})=\frac{\exp(s_k)}{\sum_{j=1}^{K}\exp(s_j)},
\label{eq:softmax}
\end{equation}
and the model minimizes the weighted cross-entropy:
\begin{equation}
\mathcal{L}=-\sum_{n} w_n\log p_\theta\big(y_n\mid\mathcal{I}_n,q_n,\mathcal{C}_n\big),
\label{eq:loss}
\end{equation}
where $n$ indexes the training examples and $w_n$ compensates for two imbalances in the mixture.
The sources differ greatly in size, so $w_n$ gives every source the same total weight.
Medical classification sources are also skewed toward frequent classes, so within such a source this weight is split equally among classes, while a VQA source counts as a single class because its candidates vary across questions.
When minimizing $\mathcal{L}$, we keep the pretrained backbone largely intact by freezing the vision encoder and applying LoRA~\cite{hu2022lora} to the vision-language connector and to all linear layers of the retained language model, including the reused query and key projections.
In contrast, the marker embeddings, the retained normalizations of the last layer, and the residual MLP are either newly introduced or small, so all of their parameters are updated directly in training.

\begin{table}[htbp]
\centering
\caption{Main results on four medical benchmarks with accuracy (Acc., \%) and mean end-to-end latency (Lat., ms). Avg. is the unweighted mean and Improvement the gain over Qwen3.5-4B. Every row was measured on one H20 at batch size one, and we \textbf{bold} the best and \underline{underline} the second best.}
\label{tab:main}
\renewcommand\tabcolsep{4pt}
\renewcommand\arraystretch{1.1}
\resizebox{\textwidth}{!}{
\begin{tabular}{l|C{1.45cm}C{1.45cm}|C{1.45cm}C{1.45cm}|C{1.45cm}C{1.45cm}|C{1.45cm}C{1.45cm}|C{1.45cm}C{1.45cm}}
\Xhline{1.2pt}
\rowcolor{medjevorange!18}
 & \multicolumn{2}{c|}{\textbf{HAM10000}} & \multicolumn{2}{c|}{\textbf{PneumoniaMNIST}} & \multicolumn{2}{c|}{\textbf{MedThinkVQA}} & \multicolumn{2}{c|}{\textbf{PMC-VQA}} & \multicolumn{2}{c}{\textbf{Avg.}} \\
\rowcolor{medjevorange!18}
\multirow{-2}{*}{\textbf{Method}} & Acc.$\uparrow$ & Lat.$\downarrow$ & Acc.$\uparrow$ & Lat.$\downarrow$ & Acc.$\uparrow$ & Lat.$\downarrow$ & Acc.$\uparrow$ & Lat.$\downarrow$ & Acc.$\uparrow$ & Lat.$\downarrow$ \\
\Xhline{1.2pt}
\rowcolor{medjevorange!7}\multicolumn{11}{c}{\textbf{\textit{General Vision-Language Models}}} \\
LLaVA-1.5-7B~\cite{liu2024llava15} & 7.41 & 151.9 & 37.50 & 140.0 & 6.81 & \underline{144.6} & 35.43 & 144.3 & 21.79 & \underline{145.2} \\
\rowcolor{gray!10}LLaVA-1.6-Vicuna-7B~\cite{liu2024llavanext} & 3.90 & 371.0 & 37.50 & 211.8 & 0.00 & \textbf{139.2} & 39.43 & 257.6 & 20.21 & 244.9 \\
LLaVA-1.6-Mistral-7B~\cite{liu2024llavanext} & 4.10 & 408.2 & 37.34 & 243.0 & 19.72 & 2,810.7 & 41.03 & 290.7 & 25.55 & 938.2 \\
\rowcolor{gray!10}LLaVA-OneVision-7B~\cite{li2024llavaonevision} & 40.48 & 435.2 & 40.38 & 234.1 & 35.14 & 909.3 & 56.22 & 251.1 & 43.06 & 457.4 \\
InternVL2.5-4B~\cite{chen2024internvl25} & 53.31 & 801.4 & 51.44 & 172.7 & 29.58 & 765.3 & 51.05 & 578.1 & 46.34 & 579.4 \\
\rowcolor{gray!10}InternVL3-9B~\cite{zhu2025internvl3} & 62.17 & 1,404.2 & 51.12 & 247.9 & 32.08 & 892.5 & 54.07 & 940.1 & 49.86 & 871.2 \\
InternVL3.5-4B~\cite{wang2025internvl35} & 59.26 & 1,001.8 & 58.33 & 106.5 & 38.47 & 836.1 & 59.06 & 532.4 & 53.78 & 619.2 \\
\rowcolor{gray!10}InternVL3.5-8B~\cite{wang2025internvl35} & \underline{65.41} & 1,032.3 & 51.28 & 132.0 & 32.50 & 750.7 & 60.66 & 677.4 & 52.46 & 648.1 \\
Keye-VL-8B-Preview~\cite{kwai2025keyevl} & 61.24 & 151.0 & 39.74 & 127.7 & 27.78 & 1,124.2 & 56.09 & 128.2 & 46.21 & 382.8 \\
\rowcolor{gray!10}Keye-VL-1.5-8B~\cite{yang2025keyevl15} & 58.07 & 160.8 & 38.46 & 120.6 & 26.67 & 615.9 & 55.15 & 130.7 & 44.59 & 257.0 \\
MiMo-VL-7B-RL~\cite{xiaomi2025mimovl} & 31.48 & 371.7 & 50.32 & 320.2 & 24.17 & 8,806.5 & 49.81 & 1,353.0 & 38.95 & 2,712.9 \\
\rowcolor{gray!10}Qwen2.5-VL-7B~\cite{bai2025qwen25vl} & 23.88 & 159.3 & 39.42 & \underline{100.3} & 32.22 & 740.1 & 53.27 & 113.6 & 37.20 & 278.4 \\
Qwen3-VL-4B~\cite{bai2025qwen3vl} & 53.17 & \underline{112.6} & 66.51 & 124.8 & 25.97 & 365.2 & 51.93 & \underline{113.1} & 49.39 & 178.9 \\
\rowcolor{gray!10}Qwen3-VL-8B~\cite{bai2025qwen3vl} & 59.66 & 135.3 & 60.10 & 106.8 & 30.42 & 489.7 & 55.46 & 128.1 & 51.41 & 215.0 \\
Qwen3.5-9B~\cite{qwen2026qwen35_9b} & 64.95 & 209.9 & 85.42 & 196.7 & \underline{40.69} & 20,947.7 & \underline{61.19} & 964.9 & \underline{63.06} & 5,579.8 \\
\hline
\rowcolor{medjevorange!7}\multicolumn{11}{c}{\textbf{\textit{Medical Vision-Language Models}}} \\
LLaVA-Med-v1.5-7B~\cite{li2023llavamed} & 0.00 & 505.5 & 18.91 & 397.1 & 0.00 & 231.7 & 1.30 & 840.4 & 5.05 & 493.7 \\
\rowcolor{gray!10}HuatuoGPT-Vision-7B~\cite{chen2024huatuogptvision} & 26.52 & 249.3 & 85.42 & 161.8 & 35.97 & 818.6 & 52.89 & 170.1 & 50.20 & 350.0 \\
MedGemma-4B~\cite{sellergren2025medgemma} & 56.48 & 215.2 & 39.10 & 207.8 & 31.94 & 949.9 & 48.44 & 213.0 & 43.99 & 396.5 \\
\rowcolor{gray!10}Lingshu-7B~\cite{xu2025lingshu} & 60.25 & 184.9 & 82.53 & 126.8 & 39.03 & 771.1 & 57.45 & 141.7 & 59.81 & 306.1 \\
\hline
Qwen3.5-4B~\cite{qwen2026qwen35} & 58.40 & 257.3 & \underline{92.79} & 267.3 & 36.39 & 1,842.8 & 56.95 & 372.7 & 61.13 & 685.0 \\
\rowcolor{gray!10}\textsc{MedJev}-4B (Ours) & \textbf{68.39} & \textbf{73.3} & \textbf{94.23} & \textbf{64.8} & \textbf{51.94} & 250.7 & \textbf{61.68} & \textbf{65.0} & \textbf{69.06} & \textbf{113.4} \\
\textbf{\textsc{Improvement}} & \textcolor{medjevorange}{+9.99} & \textcolor{medjevorange}{3.51$\times$} & \textcolor{medjevorange}{+1.44} & \textcolor{medjevorange}{4.13$\times$} & \textcolor{medjevorange}{+15.56} & \textcolor{medjevorange}{7.35$\times$} & \textcolor{medjevorange}{+4.73} & \textcolor{medjevorange}{5.74$\times$} & \textcolor{medjevorange}{+7.93} & \textcolor{medjevorange}{6.04$\times$} \\
\Xhline{1.2pt}
\end{tabular}}
\end{table}

At deployment, the LoRA updates are merged into the pretrained weights and the marker embeddings into the embedding table, so the deployed model has exactly the reduced architecture of Sec.~\ref{sec:depth} and no adapter overhead.
A prediction then takes one forward pass over the input of Eq.~\eqref{eq:input} followed by Eqs.~\eqref{eq:qk} to \eqref{eq:score} and the selection in Eq.~\eqref{eq:argmax}.
Since no token is generated, it needs no decoding loop, no cache of generated tokens, and no answer parser.
As a result, the latency depends on the input and the reduced depth alone, and every prediction carries a distribution over the candidates from Eq.~\eqref{eq:softmax}.

\section{Experiments}
\label{sec:experiments}

\subsection{Experimental Setup}
\label{sec:setup}

\subsubsection{Benchmarks}
\label{sec:benchmarks}
We conduct experiments on four medical image understanding benchmarks that differ in imaging modality and in decision type.
HAM10000~\cite{tschandl2018ham10000} asks for one of seven dermoscopic diagnosis categories and provides 1,512 test examples.
PneumoniaMNIST~\cite{yang2023medmnist} asks whether a pediatric chest radiograph is normal or shows pneumonia and provides 624 test examples.
MedThinkVQA~\cite{yao2026medicalthinking} presents a clinical case with 2 to 47 images and five candidate diagnoses and provides 720 test examples.
PMC-VQA~\cite{zhang2023pmcvqa} asks a question about a figure from the biomedical literature with four options and provides 33,430 test examples.
The four benchmarks together contain 36,286 test examples, and the number of candidate answers per example ranges from two to seven.
HAM10000 and PneumoniaMNIST label an image with a fixed set of diagnoses, while MedThinkVQA and PMC-VQA ask a question about each case, so the mixture covers classification over a fixed label set and question answering over free text.

\subsubsection{Baselines}
\label{sec:baselines}
We compare \textsc{MedJev}-4B with 20 off-the-shelf vision-language models at the 4B to 9B scale.
Fifteen of them are general-purpose models from the LLaVA~\cite{liu2024llava15,liu2024llavanext,li2024llavaonevision}, InternVL~\cite{chen2024internvl25,zhu2025internvl3,wang2025internvl35}, Keye-VL~\cite{kwai2025keyevl,yang2025keyevl15}, MiMo-VL~\cite{xiaomi2025mimovl}, and Qwen~\cite{bai2025qwen25vl,bai2025qwen3vl,qwen2026qwen35_9b} families, and four are medical models trained on biomedical image-text data~\cite{li2023llavamed,chen2024huatuogptvision,sellergren2025medgemma,xu2025lingshu}.
The remaining one is the original Qwen3.5-4B~\cite{qwen2026qwen35}, which we compare against most closely because it shares the vision encoder, the connector, and the retained language layers with \textsc{MedJev}-4B.
Every baseline runs off the shelf with its official weights and chat template, and a response naming no candidate counts as an error.
We restrict the comparison to the 4B to 9B range so that every baseline stays close to \textsc{MedJev}-4B in size.

\begin{figure}[htbp]
  \centering
  \includegraphics[width=\textwidth]{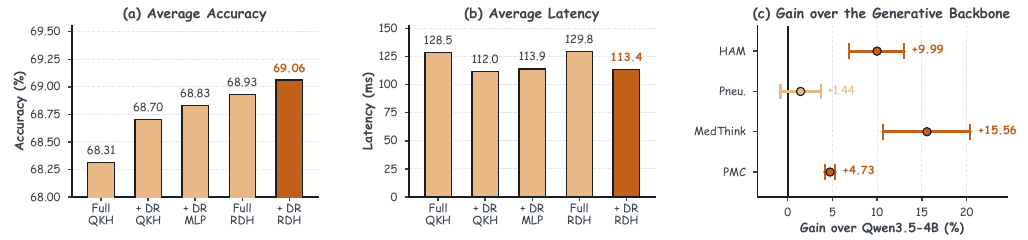}
  \caption{\textbf{Ablation and significance.} (a) Unweighted mean accuracy and (b) mean end-to-end latency over the four benchmarks, where DR is the depth reduction, QKH the query-key head, MLP the per-candidate MLP and RDH the residual decision head. (c) The accuracy gain of \textsc{MedJev}-4B over the generative Qwen3.5-4B, with its 95\% interval from a paired bootstrap of 10,000 resamples.}
  \label{fig:ablation}
\end{figure}

\subsubsection{Implementation Details}
\label{sec:implementation}
We build \textsc{MedJev}-4B on Qwen3.5-4B, whose language model has $L=32$ layers with a full-attention last layer.
We keep the first $J=23$ layers, which reduces the parameter count from 4.44B to 3.55B, and the residual MLP has a hidden dimension of 256 and adds 1.97M parameters for $D=2560$.
In Qwen3.5, the query projection also produces an output gate, which we discard.
We train a single model for one epoch on the 174,360 training examples of the four benchmarks with LoRA of rank 16 and scaling 32, which amounts to 5,449 updates with seed 42.
We measure the end-to-end latency of every model on an NVIDIA H20 GPU at batch size one, including image loading, preprocessing, and model execution.
Each baseline is timed in a separate session under the same protocol, and the latency of Qwen3.5-4B comes from the session that also times \textsc{MedJev}-4B, so that the original backbone and our model are measured under exactly the same timing conditions.

\subsection{Main Results}
\label{sec:main_results}
As shown in Table~\ref{tab:main}, \textsc{MedJev}-4B is the most accurate model on all four benchmarks and the fastest on three of them and on average, whereas no baseline reaches both ends of that trade-off, with the most accurate one spending 5,579.8\,ms on a case and the fastest one answering 21.79\% of them correctly.
The comparison that carries weight is Qwen3.5-4B, which shares the vision encoder, the connector and the first $J$ language layers with our model, and against it we gain 7.93\% of average accuracy while deciding 6.04$\times$ faster.
Removing the decoding loop produces both gains, since a generated answer spends computation on tokens the decision never uses, and it spends more the longer the answer runs.
Generative latency therefore spans two orders of magnitude across the four benchmarks, and MedThinkVQA shows it most clearly, where Qwen3.5-9B spends 20,947.7\,ms on a case to reach the second highest average accuracy and \textsc{MedJev}-4B spends 250.7\,ms.
Free text also fails in two ways that a decision cannot.
The multi-image cases of MedThinkVQA overflow the context window of LLaVA-Med-v1.5-7B and LLaVA-1.6-Vicuna-7B and leave nothing to read, and on HAM10000 the first of those two answers with the option letter inside a sentence that the fixed rule set of the protocol will not accept.
A \textsc{MedJev}-4B prediction is a candidate index, so the parsing failure cannot occur, and the fixed visual budget keeps every case here inside the input window.

\subsection{Further Analysis}
\label{sec:analysis}

\subsubsection{Ablation Study}
\label{sec:ablation}
Fig.~\ref{fig:ablation} (a) and Fig.~\ref{fig:ablation} (b) vary the retained depth and the decision head and hold everything else fixed, with the accuracy axis zoomed so that every bar carries its value.
Full keeps all 32 language layers and DR reduces them to $J=23$, QKH scores a candidate by pretrained query-key matching alone, MLP replaces that matching with a per-candidate MLP, and RDH adds the zero-initialized residual we deploy.
The depth reduction costs no accuracy under either head, raising the mean from 68.31\% to 68.70\% under QKH and from 68.93\% to 69.06\% under RDH while cutting about 16\,ms of latency.
The depth probe of Fig.~\ref{fig:depth_probe} anticipates this, since a linear decision is already readable at layer 23 and the layers above serve the vocabulary head.
The head comparison is closer.
QKH alone reaches 68.70\%, so the pretrained attention geometry carries most of the decision, and the MLP scorer adds little at 68.83\%.
RDH reaches the best mean of 69.06\% for 1.97M added parameters and 1.4\,ms over QKH, refining the matching score that MLP simply replaces.

\subsubsection{Significance of the Gains}
\label{sec:significance}
In Fig.~\ref{fig:ablation} (c), we resample the test examples 10,000 times and recompute both accuracies, which turns each gain of Table~\ref{tab:main} into a paired bootstrap interval.
Three of the four stay clear of zero, so the gains of 9.99\% on HAM10000, 15.56\% on MedThinkVQA and 4.73\% on PMC-VQA exceed the sampling noise of their test sets.
The exception is the 1.44\% on PneumoniaMNIST, spanning $[-0.80, 3.69]$, and the narrowest interval belongs to PMC-VQA, so separation follows test set size as much as gain size.
The intervals rest on disagreements and not on shared successes, since \textsc{MedJev}-4B alone is correct on 5,615 PMC-VQA questions while Qwen3.5-4B alone is correct on 4,035, and the split on MedThinkVQA is 221 against 109, so the gains rest on decisions that actually flip.

\begin{figure}[htbp]
  \centering
  \includegraphics[width=\textwidth]{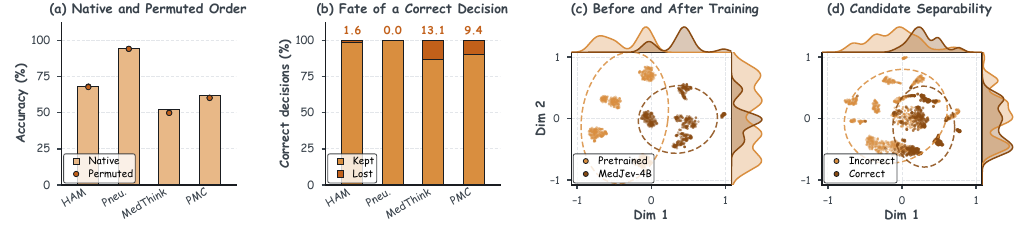}
  \caption{\textbf{Candidate order and hidden representations.} (a) Accuracy under the native and three permuted orders. (b) The fate of a correct decision, kept or lost. (c, d) t-SNE of the decision marker before and after training and of the trained candidates, with the marginal distribution of each axis.}
  \label{fig:order_repr}
\end{figure}

\subsubsection{Candidate Order}
\label{sec:order}
In Fig.~\ref{fig:order_repr} (a) and Fig.~\ref{fig:order_repr} (b), we decide every test example again with the candidate order permuted under three seeds, because a model that scores candidates could be reading the slot an option occupies instead of its text.
The permuted accuracies sit on the native bars and the unweighted mean falls only from 69.06\% to 68.10\%, so the order is not what the head reads.
Panel (b) locates the small remainder, since a fixed label vocabulary loses nothing on PneumoniaMNIST and 1.6\% on HAM10000 while free-form options lose 13.1\% on MedThinkVQA and 9.4\% on PMC-VQA.

\subsubsection{Hidden Representations}
\label{sec:representations}
In Fig.~\ref{fig:order_repr} (c) and Fig.~\ref{fig:order_repr} (d), we reduce the states the head reads on 597 training examples with t-SNE and draw the marginal distribution of both axes.
Before training the decision markers break into clusters, where two examples of one benchmark sit 2.36 times closer than two examples of different benchmarks, and training collapses them into one region at a ratio of 1.36, so one head serves candidate sets that range from two to seven answers.
The ratio compares the mean distance between two examples of one benchmark with the mean distance between two examples of different benchmarks, measured on the same sample in the reduced space.
The candidate markers separate in the same geometry, since the correct ones concentrate while the others spread out, which the matching term of Eq.~\eqref{eq:qkscore} reads, and why QKH on its own already decides well.

\needspace{22\baselineskip}
\subsubsection{Comparison with Released Decision Models}
\label{sec:jev}
\begin{wraptable}{r}{0.50\textwidth}
\vspace{-\intextsep}
\centering
\caption{\textbf{Comparison with released decision models.} Accuracy (\%) on the four benchmarks, Avg. the unweighted mean and Lat. the latency (ms) over the three single-image benchmarks.}
\label{tab:jev}
\renewcommand\tabcolsep{3.1pt}
\renewcommand\arraystretch{1.1}
\resizebox{0.48\textwidth}{!}{
\begin{tabular}{l|cccc|cc}
\Xhline{1.2pt}
\rowcolor{medjevorange!18}
\textbf{Model} & HAM & Pneu. & MedThink & PMC & Avg.$\uparrow$ & Lat.$\downarrow$ \\
\Xhline{1.2pt}
d1-3B~\cite{liquid2026d1} & 59.92 & 37.50 & 31.53 & 55.43 & 46.09 & \textbf{35.6} \\
\rowcolor{gray!10}Visual Jev-4B~\cite{yu2026visualjev} & 56.81 & 62.02 & 20.56 & 51.99 & 47.84 & 78.5 \\
OmniMed-Jev-4B~\cite{lytang2026omnimedjev} & 66.67 & \textbf{95.03} & 22.22 & 32.93 & 54.21 & 201.4 \\
\rowcolor{gray!10}PathJev-P~\cite{tommasino2026pathjev} & 51.32 & 64.10 & 26.81 & 37.23 & 44.86 & 192.1 \\
Clef-Flash-9B~\cite{cloudflare2026clef} & 61.90 & 79.97 & 28.06 & 56.37 & 56.57 & 83.1 \\
\rowcolor{gray!10}d3-flash-9B~\cite{vllmsr2026d3} & \underline{67.00} & 71.47 & \underline{32.92} & \underline{59.88} & \underline{57.82} & 75.9 \\
\textsc{MedJev}-4B (Ours) & \textbf{68.39} & \underline{94.23} & \textbf{51.81} & \textbf{61.66} & \textbf{69.02} & \underline{56.8} \\
\Xhline{1.2pt}
\end{tabular}}
\end{wraptable}

In Table~\ref{tab:jev}, we run six released system one models through the entry point each release documents, two medical and four general-purpose, and give each the candidate texts the benchmark defines.
Every model is scored from the first image of each case, which keeps the input identical across interfaces, and the latency column covers the three single-image benchmarks, as in the main table.
\textsc{MedJev}-4B is the most accurate on three of the four and leads the mean by 11.20\%, and only the 3B model is faster.
Scale does not explain the ordering, since the two 9B models reach 56.57\% and 57.82\%.
The medical pair is the narrowest, since OmniMed-Jev-4B takes PneumoniaMNIST by 0.80\% but falls to 32.93\% on PMC-VQA, while PathJev-P averages only 44.86\%, below the general-purpose d3-flash-9B everywhere.

\subsubsection{Case Study}
\label{sec:case}
\begin{wrapfigure}{r}{0.52\textwidth}
\vspace{-\intextsep}
\centering
\includegraphics[width=0.50\textwidth]{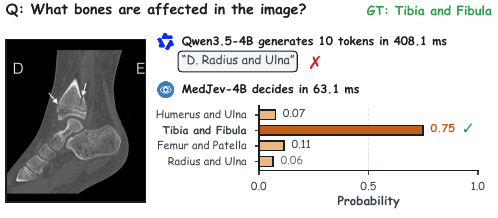}
\caption{\textbf{One PMC-VQA case under both interfaces.} The backbone decodes a sentence that names the wrong bones, while \textsc{MedJev}-4B scores all four candidates in one pass and picks the reference.}
\label{fig:casestudy}
\vspace{-\intextsep}
\end{wrapfigure}
In Fig.~\ref{fig:casestudy}, we answer one PMC-VQA case with both interfaces, the backbone above and the deployed head below.
Qwen3.5-4B spends 408.1\,ms decoding ten tokens and names the bones of the forearm for an image of an ankle, while \textsc{MedJev}-4B scores the same candidates in 63.1\,ms and puts 0.75 on the reference.
The candidate the sentence names is the one the head ranks last at 0.06, and the largest share of what remains goes to the other pair of leg bones at 0.11.
A sentence commits to one of these candidates without saying how close the others were, and that margin is exactly what a single forward pass returns for free.
The decision is also 6.47 times faster than the sentence here, close to the 6.04 times that the four benchmarks average.

\section{Conclusion}
\label{sec:conclusion}
In this paper, we presented \textsc{MedJev}, a system one vision-language decision model for fast and effective medical image understanding.
Built on Qwen3.5-4B, \textsc{MedJev}-4B further employs language-side depth reduction and a residual decision head for low latency and accurate candidate scoring.
Across extensive experiments on four medical benchmarks, the effectiveness and the efficiency of \textsc{MedJev}-4B are fully validated.
In the future, we plan to extend our work to more general scenarios such as general vision-language decision models and complex visual perception tasks.

\bibliography{main}

\begin{thebibliography}{59}
\providecommand{\natexlab}[1]{#1}
\providecommand{\url}[1]{\texttt{#1}}
\expandafter\ifx\csname urlstyle\endcsname\relax
  \providecommand{\doi}[1]{doi: #1}\else
  \providecommand{\doi}{doi: \begingroup \urlstyle{rm}\Url}\fi

\bibitem[Ainslie et~al.(2023)Ainslie, Lee-Thorp, Jong, Zemlyanskiy, Lebrón, and Sanghai]{ainslie2023gqa}
Joshua Ainslie, James Lee-Thorp, Michiel~de Jong, Yury Zemlyanskiy, Federico Lebrón, and Sumit Sanghai.
\newblock {GQA}: Training generalized multi-query transformer models from multi-head checkpoints.
\newblock In \emph{Conference on Empirical Methods in Natural Language Processing (EMNLP)}, 2023.
\newblock URL \url{https://arxiv.org/abs/2305.13245}.

\bibitem[Almeida(2026)]{almeida2026jev}
Diogo Almeida.
\newblock Introducing {System One} models \& {Jev}.
\newblock TypeSafe AI Blog, 2026.
\newblock URL \url{https://typesafe.ai/blog/introducing-system-one-models-and-jev}.
\newblock Published September 15, 2026.

\bibitem[Bai et~al.(2025{\natexlab{a}})Bai, Cai, Chen, Chen, Chen, Cheng, Deng, Ding, Gao, Ge, et~al.]{bai2025qwen3vl}
Shuai Bai, Yuxuan Cai, Ruizhe Chen, Keqin Chen, Xionghui Chen, Zesen Cheng, Lianghao Deng, Wei Ding, Chang Gao, Chunjiang Ge, et~al.
\newblock {Qwen3-VL} technical report.
\newblock \emph{arXiv preprint arXiv:2511.21631}, 2025{\natexlab{a}}.
\newblock URL \url{https://arxiv.org/abs/2511.21631}.

\bibitem[Bai et~al.(2025{\natexlab{b}})Bai, Chen, Liu, Wang, Ge, Song, Dang, Wang, Wang, Tang, et~al.]{bai2025qwen25vl}
Shuai Bai, Keqin Chen, Xuejing Liu, Jialin Wang, Wenbin Ge, Sibo Song, Kai Dang, Peng Wang, Shijie Wang, Jun Tang, et~al.
\newblock {Qwen2.5-VL} technical report.
\newblock \emph{arXiv preprint arXiv:2502.13923}, 2025{\natexlab{b}}.
\newblock URL \url{https://arxiv.org/abs/2502.13923}.

\bibitem[Chen et~al.(2024{\natexlab{a}})Chen, Gui, Ouyang, Gao, Chen, Chen, Wang, Zhang, Cai, Ji, Yu, Wan, and Wang]{chen2024huatuogptvision}
Junying Chen, Chi Gui, Ruyi Ouyang, Anningzhe Gao, Shunian Chen, Guiming~Hardy Chen, Xidong Wang, Ruifei Zhang, Zhenyang Cai, Ke~Ji, Guangjun Yu, Xiang Wan, and Benyou Wang.
\newblock {HuatuoGPT-Vision}, towards injecting medical visual knowledge into multimodal {LLMs} at scale.
\newblock \emph{arXiv preprint arXiv:2406.19280}, 2024{\natexlab{a}}.
\newblock URL \url{https://arxiv.org/abs/2406.19280}.

\bibitem[Chen et~al.(2024{\natexlab{b}})Chen, Zhao, Liu, Bai, Lin, Zhou, and Chang]{chen2024fastv}
Liang Chen, Haozhe Zhao, Tianyu Liu, Shuai Bai, Junyang Lin, Chang Zhou, and Baobao Chang.
\newblock An image is worth 1/2 tokens after layer 2: Plug-and-play inference acceleration for large vision-language models.
\newblock In \emph{European Conference on Computer Vision (ECCV)}, 2024{\natexlab{b}}.
\newblock URL \url{https://arxiv.org/abs/2403.06764}.

\bibitem[Chen et~al.(2024{\natexlab{c}})Chen, Wang, Cao, Liu, et~al.]{chen2024internvl25}
Zhe Chen, Weiyun Wang, Yue Cao, Yangzhou Liu, et~al.
\newblock Expanding performance boundaries of open-source multimodal models with model, data, and test-time scaling.
\newblock \emph{arXiv preprint arXiv:2412.05271}, 2024{\natexlab{c}}.
\newblock URL \url{https://arxiv.org/abs/2412.05271}.

\bibitem[{Cloudflare}(2026)]{cloudflare2026clef}
{Cloudflare}.
\newblock Introducing {Clef}: Our open-source decision models, and new {RL} fine-tuning platform.
\newblock Blog post, 2026.
\newblock URL \url{https://blog.cloudflare.com/clef-decision-models/}.

\bibitem[Evans and Stanovich(2013)]{evans2013dualprocess}
Jonathan St. B.~T. Evans and Keith~E. Stanovich.
\newblock Dual-process theories of higher cognition: Advancing the debate.
\newblock \emph{Perspectives on Psychological Science}, 8\penalty0 (3):\penalty0 223--241, 2013.
\newblock \doi{10.1177/1745691612460685}.
\newblock URL \url{https://doi.org/10.1177/1745691612460685}.

\bibitem[Gromov et~al.(2025)Gromov, Tirumala, Shapourian, Glorioso, and Roberts]{gromov2025unreasonable}
Andrey Gromov, Kushal Tirumala, Hassan Shapourian, Paolo Glorioso, and Daniel~A. Roberts.
\newblock The unreasonable ineffectiveness of the deeper layers.
\newblock In \emph{International Conference on Learning Representations (ICLR)}, 2025.
\newblock URL \url{https://arxiv.org/abs/2403.17887}.

\bibitem[Guo et~al.(2025)Guo, Yang, Zhang, et~al.]{deepseek2025r1}
Daya Guo, Dejian Yang, Haowei Zhang, et~al.
\newblock {DeepSeek-R1} incentivizes reasoning in {LLMs} through reinforcement learning.
\newblock \emph{Nature}, 645\penalty0 (8081):\penalty0 633--638, 2025.
\newblock \doi{10.1038/s41586-025-09422-z}.
\newblock URL \url{https://www.nature.com/articles/s41586-025-09422-z}.

\bibitem[He et~al.(2020)He, Zhang, Mou, Xing, and Xie]{he2020pathvqa}
Xuehai He, Yichen Zhang, Luntian Mou, Eric Xing, and Pengtao Xie.
\newblock {PathVQA}: 30000+ questions for medical visual question answering.
\newblock \emph{arXiv preprint arXiv:2003.10286}, 2020.
\newblock URL \url{https://arxiv.org/abs/2003.10286}.

\bibitem[Hu et~al.(2022)Hu, Shen, Wallis, Allen-Zhu, Li, Wang, Wang, and Chen]{hu2022lora}
Edward~J. Hu, Yelong Shen, Phillip Wallis, Zeyuan Allen-Zhu, Yuanzhi Li, Shean Wang, Lu~Wang, and Weizhu Chen.
\newblock {LoRA}: Low-rank adaptation of large language models.
\newblock In \emph{International Conference on Learning Representations (ICLR)}, 2022.
\newblock URL \url{https://arxiv.org/abs/2106.09685}.

\bibitem[Hu et~al.(2024)Hu, Li, Lu, Shao, He, Qiao, and Luo]{hu2024omnimedvqa}
Yutao Hu, Tianbin Li, Quanfeng Lu, Wenqi Shao, Junjun He, Yu~Qiao, and Ping Luo.
\newblock {OmniMedVQA}: A new large-scale comprehensive evaluation benchmark for medical {LVLM}.
\newblock \emph{arXiv preprint arXiv:2402.09181}, 2024.
\newblock URL \url{https://arxiv.org/abs/2402.09181}.

\bibitem[Kahneman(2011)]{kahneman2011thinking}
Daniel Kahneman.
\newblock \emph{Thinking, Fast and Slow}.
\newblock Farrar, Straus and Giroux, New York, 2011.

\bibitem[{Kwai Keye Team}(2025)]{kwai2025keyevl}
{Kwai Keye Team}.
\newblock Kwai {Keye-VL} technical report.
\newblock \emph{arXiv preprint arXiv:2507.01949}, 2025.
\newblock URL \url{https://arxiv.org/abs/2507.01949}.

\bibitem[Lai et~al.(2025)Lai, Zhong, Li, Zhao, Li, Psounis, and Yang]{lai2025medr1}
Yuxiang Lai, Jike Zhong, Ming Li, Shitian Zhao, Yuheng Li, Konstantinos Psounis, and Xiaofeng Yang.
\newblock {Med-R1}: Reinforcement learning for generalizable medical reasoning in vision-language models.
\newblock \emph{arXiv preprint arXiv:2503.13939}, 2025.
\newblock URL \url{https://arxiv.org/abs/2503.13939}.

\bibitem[Lau et~al.(2018)Lau, Gayen, Ben~Abacha, and Demner-Fushman]{lau2018vqarad}
Jason~J. Lau, Soumya Gayen, Asma Ben~Abacha, and Dina Demner-Fushman.
\newblock A dataset of clinically generated visual questions and answers about radiology images.
\newblock \emph{Scientific Data}, 5:\penalty0 180251, 2018.
\newblock \doi{10.1038/sdata.2018.251}.

\bibitem[Li et~al.(2024)Li, Zhang, Guo, Zhang, Li, Zhang, Zhang, Zhang, Li, Liu, et~al.]{li2024llavaonevision}
Bo~Li, Yuanhan Zhang, Dong Guo, Renrui Zhang, Feng Li, Hao Zhang, Kaichen Zhang, Peiyuan Zhang, Yanwei Li, Ziwei Liu, et~al.
\newblock {LLaVA-OneVision}: Easy visual task transfer.
\newblock \emph{arXiv preprint arXiv:2408.03326}, 2024.
\newblock URL \url{https://arxiv.org/abs/2408.03326}.

\bibitem[Li et~al.(2023)Li, Wong, Zhang, Usuyama, Liu, Yang, Naumann, Poon, and Gao]{li2023llavamed}
Chunyuan Li, Cliff Wong, Sheng Zhang, Naoto Usuyama, Haotian Liu, Jianwei Yang, Tristan Naumann, Hoifung Poon, and Jianfeng Gao.
\newblock {LLaVA-Med}: Training a large language-and-vision assistant for biomedicine in one day.
\newblock \emph{arXiv preprint arXiv:2306.00890}, 2023.
\newblock URL \url{https://arxiv.org/abs/2306.00890}.

\bibitem[{Liquid AI}(2026)]{liquid2026d1}
{Liquid AI}.
\newblock Open {d1}: Edge decision models for text, vision, and audio.
\newblock Blog post, 2026.
\newblock URL \url{https://www.liquid.ai/blog/d1-open}.

\bibitem[Liu et~al.(2021)Liu, Zhan, Xu, Ma, Yang, and Wu]{liu2021slake}
Bo~Liu, Li-Ming Zhan, Li~Xu, Lin Ma, Yan Yang, and Xiao-Ming Wu.
\newblock {SLAKE}: A semantically-labeled knowledge-enhanced dataset for medical visual question answering.
\newblock In \emph{IEEE International Symposium on Biomedical Imaging (ISBI)}, 2021.
\newblock URL \url{https://arxiv.org/abs/2102.09542}.

\bibitem[Liu et~al.(2026)Liu, Wang, Wang, Cheng, and Fu]{liu2026retijev}
Bo~Liu, Lianyu Wang, Meng Wang, Ching-Yu Cheng, and Huazhu Fu.
\newblock {RetiJev}: Fast and source-grounded ophthalmic decision models.
\newblock \emph{arXiv preprint arXiv:2610.30947}, 2026.
\newblock URL \url{https://arxiv.org/abs/2610.30947}.

\bibitem[Liu et~al.(2023)Liu, Li, Wu, and Lee]{liu2023llava}
Haotian Liu, Chunyuan Li, Qingyang Wu, and Yong~Jae Lee.
\newblock Visual instruction tuning.
\newblock In \emph{Advances in Neural Information Processing Systems (NeurIPS)}, 2023.
\newblock URL \url{https://arxiv.org/abs/2304.08485}.

\bibitem[Liu et~al.(2024{\natexlab{a}})Liu, Li, Li, and Lee]{liu2024llava15}
Haotian Liu, Chunyuan Li, Yuheng Li, and Yong~Jae Lee.
\newblock Improved baselines with visual instruction tuning.
\newblock In \emph{IEEE/CVF Conference on Computer Vision and Pattern Recognition (CVPR)}, 2024{\natexlab{a}}.
\newblock URL \url{https://arxiv.org/abs/2310.03744}.

\bibitem[Liu et~al.(2024{\natexlab{b}})Liu, Li, Li, Li, Zhang, Shen, and Lee]{liu2024llavanext}
Haotian Liu, Chunyuan Li, Yuheng Li, Bo~Li, Yuanhan Zhang, Sheng Shen, and Yong~Jae Lee.
\newblock {LLaVA-NeXT}: Improved reasoning, {OCR}, and world knowledge.
\newblock Blog post, 2024{\natexlab{b}}.
\newblock URL \url{https://llava-vl.github.io/blog/2024-01-30-llava-next/}.

\bibitem[Men et~al.(2024)Men, Xu, Zhang, Wang, Lin, Lu, Han, and Chen]{men2024shortgpt}
Xin Men, Mingyu Xu, Qingyu Zhang, Bingning Wang, Hongyu Lin, Yaojie Lu, Xianpei Han, and Weipeng Chen.
\newblock {ShortGPT}: Layers in large language models are more redundant than you expect.
\newblock \emph{arXiv preprint arXiv:2403.03853}, 2024.
\newblock URL \url{https://arxiv.org/abs/2403.03853}.

\bibitem[Moor et~al.(2023)Moor, Huang, Wu, Yasunaga, Zakka, Dalmia, Reis, Rajpurkar, and Leskovec]{moor2023medflamingo}
Michael Moor, Qian Huang, Shirley Wu, Michihiro Yasunaga, Cyril Zakka, Yash Dalmia, Eduardo~Pontes Reis, Pranav Rajpurkar, and Jure Leskovec.
\newblock {Med-Flamingo}: A multimodal medical few-shot learner.
\newblock \emph{arXiv preprint arXiv:2307.15189}, 2023.
\newblock URL \url{https://arxiv.org/abs/2307.15189}.

\bibitem[Pan et~al.(2025)Pan, Liu, Wu, Liu, Zhu, Li, Chen, Ouyang, and Rueckert]{pan2025medvlmr1}
Jiazhen Pan, Che Liu, Junde Wu, Fenglin Liu, Jiayuan Zhu, Hongwei~Bran Li, Chen Chen, Cheng Ouyang, and Daniel Rueckert.
\newblock {MedVLM-R1}: Incentivizing medical reasoning capability of vision-language models (vlms) via reinforcement learning.
\newblock \emph{arXiv preprint arXiv:2502.19634}, 2025.
\newblock URL \url{https://arxiv.org/abs/2502.19634}.

\bibitem[Qin et~al.(2026)Qin, Zhu, Zhou, Qiao, Gao, Zhou, Zhang, Zeng, Liu, Lu, Chen, Chen, and Ma]{qin2026mjevspatial}
Luozheng Qin, Yuhan Zhu, Changhai Zhou, Qian Qiao, Jun Gao, Yuhua Zhou, Harry Zhang, Xiangyu Zeng, Yuxuan Liu, Yuxin Lu, Kaijie Chen, Andrew Chen, and Pony Ma.
\newblock {MJev-Spatial}: {System One} spatial intelligence via native attention matching.
\newblock \emph{arXiv preprint arXiv:2610.22276}, 2026.
\newblock URL \url{https://arxiv.org/abs/2610.22276}.

\bibitem[{Qwen Team}(2026{\natexlab{a}})]{qwen2026qwen35}
{Qwen Team}.
\newblock {Qwen3.5-4B}: Official model card.
\newblock Hugging Face, 2026{\natexlab{a}}.
\newblock URL \url{https://huggingface.co/Qwen/Qwen3.5-4B}.

\bibitem[{Qwen Team}(2026{\natexlab{b}})]{qwen2026qwen35_9b}
{Qwen Team}.
\newblock {Qwen3.5-9B}: Official model card.
\newblock Hugging Face, 2026{\natexlab{b}}.
\newblock URL \url{https://huggingface.co/Qwen/Qwen3.5-9B}.

\bibitem[Robinson et~al.(2023)Robinson, Rytting, and Wingate]{robinson2023mcqa}
Joshua Robinson, Christopher~Michael Rytting, and David Wingate.
\newblock Leveraging large language models for multiple choice question answering.
\newblock In \emph{International Conference on Learning Representations (ICLR)}, 2023.
\newblock URL \url{https://arxiv.org/abs/2210.12353}.

\bibitem[Sellergren et~al.(2025)Sellergren, Kazemzadeh, Jaroensri, Kiraly, Traverse, Kohlberger, Xu, Jamil, Hughes, Lau, et~al.]{sellergren2025medgemma}
Andrew Sellergren, Sahar Kazemzadeh, Tiam Jaroensri, Atilla Kiraly, Madeleine Traverse, Timo Kohlberger, Shawn Xu, Fayaz Jamil, C{\'\i}an Hughes, Charles Lau, et~al.
\newblock {MedGemma} technical report.
\newblock \emph{arXiv preprint arXiv:2507.05201}, 2025.
\newblock URL \url{https://arxiv.org/abs/2507.05201}.

\bibitem[Shang et~al.(2025)Shang, Cai, Xu, Lee, and Yan]{shang2025prumerge}
Yuzhang Shang, Mu~Cai, Bingxin Xu, Yong~Jae Lee, and Yan Yan.
\newblock {LLaVA-PruMerge}: Adaptive token reduction for efficient large multimodal models.
\newblock In \emph{IEEE/CVF International Conference on Computer Vision (ICCV)}, 2025.
\newblock URL \url{https://arxiv.org/abs/2403.15388}.

\bibitem[Su et~al.(2021)Su, Lu, Pan, Murtadha, Wen, and Liu]{su2021roformer}
Jianlin Su, Yu~Lu, Shengfeng Pan, Ahmed Murtadha, Bo~Wen, and Yunfeng Liu.
\newblock {RoFormer}: Enhanced transformer with rotary position embedding.
\newblock \emph{arXiv preprint arXiv:2104.09864}, 2021.
\newblock URL \url{https://arxiv.org/abs/2104.09864}.

\bibitem[Tang and Chen(2026)]{lytang2026omnimedjev}
Luyao Tang and Cheng Chen.
\newblock {OmniMed-Jev}: Calibrating {LVLM} confidence for trustworthy medical multimodal decisions via system one.
\newblock \emph{arXiv preprint arXiv:2610.00381}, 2026.
\newblock URL \url{https://arxiv.org/abs/2610.00381}.

\bibitem[Tommasino and Rinaldi(2026)]{tommasino2026pathjev}
Cristian Tommasino and Antonio~M. Rinaldi.
\newblock {PathJev}: Decision-native vision-language modelling for histopathology.
\newblock \emph{arXiv preprint arXiv:2610.32478}, 2026.
\newblock URL \url{https://arxiv.org/abs/2610.32478}.

\bibitem[Tschandl et~al.(2018)Tschandl, Rosendahl, and Kittler]{tschandl2018ham10000}
Philipp Tschandl, Cliff Rosendahl, and Harald Kittler.
\newblock The {HAM10000} dataset, a large collection of multi-source dermatoscopic images of common pigmented skin lesions.
\newblock \emph{Scientific Data}, 5:\penalty0 180161, 2018.
\newblock \doi{10.1038/sdata.2018.161}.
\newblock URL \url{https://www.nature.com/articles/sdata2018161}.

\bibitem[Tu et~al.(2023)Tu, Azizi, Driess, Schaekermann, Amin, Chang, Carroll, Lau, Tanno, Ktena, et~al.]{tu2023medpalmm}
Tao Tu, Shekoofeh Azizi, Danny Driess, Mike Schaekermann, Mohamed Amin, Pi-Chuan Chang, Andrew Carroll, Chuck Lau, Ryutaro Tanno, Ira Ktena, et~al.
\newblock Towards generalist biomedical {AI}.
\newblock \emph{arXiv preprint arXiv:2307.14334}, 2023.
\newblock URL \url{https://arxiv.org/abs/2307.14334}.

\bibitem[van~der Maaten and Hinton(2008)]{maaten2008tsne}
Laurens van~der Maaten and Geoffrey Hinton.
\newblock Visualizing data using {t-SNE}.
\newblock \emph{Journal of Machine Learning Research}, 9\penalty0 (86):\penalty0 2579--2605, 2008.
\newblock URL \url{https://jmlr.org/papers/v9/vandermaaten08a.html}.

\bibitem[{vLLM Semantic Router Team}(2026)]{vllmsr2026d3}
{vLLM Semantic Router Team}.
\newblock {d3-flash}: A multimodal foundation decision model.
\newblock Model release, 2026.
\newblock URL \url{https://huggingface.co/vllm-sr/d3-flash}.

\bibitem[Wang et~al.(2025)Wang, Gao, Gu, Pu, Cui, Wei, Liu, Jing, Ye, Shao, et~al.]{wang2025internvl35}
Weiyun Wang, Zhangwei Gao, Lixin Gu, Hengjun Pu, Long Cui, Xingguang Wei, Zhaoyang Liu, Linglin Jing, Shenglong Ye, Jie Shao, et~al.
\newblock {InternVL3.5}: Advancing open-source multimodal models in versatility, reasoning, and efficiency.
\newblock \emph{arXiv preprint arXiv:2508.18265}, 2025.
\newblock URL \url{https://arxiv.org/abs/2508.18265}.

\bibitem[Wei et~al.(2022)Wei, Wang, Schuurmans, Bosma, Ichter, Xia, Chi, Le, and Zhou]{wei2022cot}
Jason Wei, Xuezhi Wang, Dale Schuurmans, Maarten Bosma, Brian Ichter, Fei Xia, Ed~Chi, Quoc Le, and Denny Zhou.
\newblock Chain-of-thought prompting elicits reasoning in large language models.
\newblock In \emph{Advances in Neural Information Processing Systems (NeurIPS)}, 2022.
\newblock URL \url{https://arxiv.org/abs/2201.11903}.

\bibitem[Wu et~al.(2023)Wu, Zhang, Zhang, Wang, and Xie]{wu2023radfm}
Chaoyi Wu, Xiaoman Zhang, Ya~Zhang, Yanfeng Wang, and Weidi Xie.
\newblock Towards generalist foundation model for radiology by leveraging web-scale {2D\&3D} medical data.
\newblock \emph{arXiv preprint arXiv:2308.02463}, 2023.
\newblock URL \url{https://arxiv.org/abs/2308.02463}.

\bibitem[{Xiaomi LLM-Core Team}(2025)]{xiaomi2025mimovl}
{Xiaomi LLM-Core Team}.
\newblock {MiMo-VL} technical report.
\newblock \emph{arXiv preprint arXiv:2506.03569}, 2025.
\newblock URL \url{https://arxiv.org/abs/2506.03569}.

\bibitem[Xing et~al.(2025)Xing, Huang, Dong, Lu, Zhang, Zang, Cao, He, Wang, Wu, et~al.]{xing2025pyramiddrop}
Long Xing, Qidong Huang, Xiaoyi Dong, Jiajie Lu, Pan Zhang, Yuhang Zang, Yuhang Cao, Conghui He, Jiaqi Wang, Feng Wu, et~al.
\newblock Conical visual concentration for efficient large vision-language models.
\newblock In \emph{IEEE/CVF Conference on Computer Vision and Pattern Recognition (CVPR)}, 2025.
\newblock URL \url{https://arxiv.org/abs/2410.17247}.

\bibitem[Xu et~al.(2025)Xu, Chan, Li, Aljunied, Yuan, Wang, Xiao, Chen, Liu, Li, Sun, Shen, Wang, Tan, Zhao, Xu, Zhang, and Rong]{xu2025lingshu}
Weiwen Xu, Hou~Pong Chan, Long Li, Mahani Aljunied, Ruifeng Yuan, Jianyu Wang, Chenghao Xiao, Guizhen Chen, Chaoqun Liu, Zhaodonghui Li, Yu~Sun, Junao Shen, Chaojun Wang, Jie Tan, Deli Zhao, Tingyang Xu, Hao Zhang, and Yu~Rong.
\newblock Lingshu: A generalist foundation model for unified multimodal medical understanding and reasoning.
\newblock \emph{arXiv preprint arXiv:2506.07044}, 2025.
\newblock URL \url{https://arxiv.org/abs/2506.07044}.

\bibitem[Yang et~al.(2025)Yang, Wen, Ding, Liu, et~al.]{yang2025keyevl15}
Biao Yang, Bin Wen, Boyang Ding, Changyi Liu, et~al.
\newblock Kwai {Keye-VL} 1.5 technical report.
\newblock \emph{arXiv preprint arXiv:2509.01563}, 2025.
\newblock URL \url{https://arxiv.org/abs/2509.01563}.

\bibitem[Yang et~al.(2023)Yang, Shi, Wei, Liu, Zhao, Ke, Pfister, and Ni]{yang2023medmnist}
Jiancheng Yang, Rui Shi, Donglai Wei, Zequan Liu, Lin Zhao, Bilian Ke, Hanspeter Pfister, and Bingbing Ni.
\newblock {MedMNIST v2}---a large-scale lightweight benchmark for {2D} and {3D} biomedical image classification.
\newblock \emph{Scientific Data}, 2023.
\newblock \doi{10.1038/s41597-022-01721-8}.
\newblock URL \url{https://arxiv.org/abs/2110.14795}.

\bibitem[Yang et~al.(2024)Yang, Chen, Tian, Wang, Li, Yu, and Jia]{yang2024visionzip}
Senqiao Yang, Yukang Chen, Zhuotao Tian, Chengyao Wang, Jingyao Li, Bei Yu, and Jiaya Jia.
\newblock {VisionZip}: Longer is better but not necessary in vision language models.
\newblock \emph{arXiv preprint arXiv:2412.04467}, 2024.
\newblock URL \url{https://arxiv.org/abs/2412.04467}.

\bibitem[Yao et~al.(2026)Yao, Wang, Zhang, Wang, Xia, Tang, Han, Ouyang, Yang, Cohan, and Yu]{yao2026medicalthinking}
Zonghai Yao, Benlu Wang, Yifan Zhang, Junda Wang, Iris Xia, Zhipeng Tang, Shuo Han, Feiyun Ouyang, Zhichao Yang, Arman Cohan, and Hong Yu.
\newblock Medical thinking with multiple images.
\newblock In \emph{International Conference on Learning Representations (ICLR)}, 2026.
\newblock URL \url{https://arxiv.org/abs/2604.16506}.

\bibitem[Yu and Yao(2026)]{yu2026visualjev}
Guanxu Yu and Yuhang Yao.
\newblock {Visual Jev}: Accurate and efficient decisions from shared visual context.
\newblock \emph{arXiv preprint arXiv:2609.25845}, 2026.
\newblock URL \url{https://arxiv.org/abs/2609.25845}.

\bibitem[Zhang et~al.(2023{\natexlab{a}})Zhang, Xu, Usuyama, Xu, Bagga, Tinn, Preston, Rao, Wei, Valluri, et~al.]{zhang2023biomedclip}
Sheng Zhang, Yanbo Xu, Naoto Usuyama, Hanwen Xu, Jaspreet Bagga, Robert Tinn, Sam Preston, Rajesh Rao, Mu~Wei, Naveen Valluri, et~al.
\newblock {BiomedCLIP}: a multimodal biomedical foundation model pretrained from fifteen million scientific image-text pairs.
\newblock \emph{arXiv preprint arXiv:2303.00915}, 2023{\natexlab{a}}.
\newblock URL \url{https://arxiv.org/abs/2303.00915}.

\bibitem[Zhang et~al.(2023{\natexlab{b}})Zhang, Wu, Zhao, Lin, Zhang, Wang, and Xie]{zhang2023pmcvqa}
Xiaoman Zhang, Chaoyi Wu, Ziheng Zhao, Weixiong Lin, Ya~Zhang, Yanfeng Wang, and Weidi Xie.
\newblock {PMC-VQA}: Visual instruction tuning for medical visual question answering.
\newblock \emph{arXiv preprint arXiv:2305.10415}, 2023{\natexlab{b}}.
\newblock URL \url{https://arxiv.org/abs/2305.10415}.

\bibitem[Zhang et~al.(2025)Zhang, Fan, Ma, Zheng, Huang, Cheng, Gudovskiy, Okuno, Nakata, Keutzer, and Zhang]{zhang2025sparsevlm}
Yuan Zhang, Chun-Kai Fan, Junpeng Ma, Wenzhao Zheng, Tao Huang, Kuan Cheng, Denis Gudovskiy, Tomoyuki Okuno, Yohei Nakata, Kurt Keutzer, and Shanghang Zhang.
\newblock {SparseVLM}: Visual token sparsification for efficient vision-language model inference.
\newblock In \emph{International Conference on Machine Learning (ICML)}, 2025.
\newblock URL \url{https://arxiv.org/abs/2410.04417}.

\bibitem[Zhou et~al.(2026)Zhou, Yang, and Zhao]{zhou2026pixeljev}
Xunlan Zhou, Xianliang Yang, and Li~Zhao.
\newblock From text decisions to pixels: An study of {Jev}-style visual choice model.
\newblock \emph{arXiv preprint arXiv:2609.29283}, 2026.
\newblock URL \url{https://arxiv.org/abs/2609.29283}.

\bibitem[Zhu et~al.(2025)Zhu, Wang, Chen, Liu, et~al.]{zhu2025internvl3}
Jinguo Zhu, Weiyun Wang, Zhe Chen, Zhaoyang Liu, et~al.
\newblock {InternVL3}: Exploring advanced training and test-time recipes for open-source multimodal models.
\newblock \emph{arXiv preprint arXiv:2504.10479}, 2025.
\newblock URL \url{https://arxiv.org/abs/2504.10479}.

\bibitem[Zhu et~al.(2026)Zhu, Zeng, Si, Yang, Li, Qin, Zhou, Yao, Zhang, Qin, Zhang, Yang, Xu, Zhang, Guo, Wang, Duan, and Wang]{zhu2026videojev}
Yuhan Zhu, Xiangyu Zeng, Qingyi Si, Chenxu Yang, Yifei Li, Luozheng Qin, Junhao Zhou, Dingyu Yao, Jun Zhang, Chuanyu Qin, Zhiqiu Zhang, Yuandong Yang, Yuqi Xu, Tianyu Zhang, Hongcheng Guo, Jiaqi Wang, Nan Duan, and Limin Wang.
\newblock {VideoJev}: {System One} model for video understanding and interaction.
\newblock \emph{arXiv preprint arXiv:2610.18615}, 2026.
\newblock URL \url{https://arxiv.org/abs/2610.18615}.

\end{thebibliography}

\clearpage
\appendix
\setcounter{figure}{0}\renewcommand{\thefigure}{A\arabic{figure}}
\setcounter{table}{0}\renewcommand{\thetable}{A\arabic{table}}
\section{Introduction to the Datasets}
\label{supp:datasets}

We evaluate on four medical image understanding benchmarks. They differ in imaging modality, in the
number of images a case carries, and in whether the candidates are a fixed label set or free text, and
the number of candidates per example ranges from two to seven.

\paragraph{HAM10000~\cite{tschandl2018ham10000}.}
Dermoscopic images of pigmented skin lesions, each labelled with one of seven diagnostic categories.
The categories are melanoma, melanocytic nevus, basal cell carcinoma, actinic keratosis or
intraepithelial carcinoma, benign keratosis, dermatofibroma and vascular lesion. We use the dermoscopic images as the training
pool and the ISIC 2018 Task 3 split as the labelled test set, which holds 1,512 cases. The categories
are imbalanced, with melanocytic nevus covering most cases.

\paragraph{PneumoniaMNIST~\cite{yang2023medmnist}.}
Pediatric chest radiographs at $224 \times 224$ from MedMNIST, to be classified as normal or
pneumonia. We use the official split, which holds 4,708 training and 624 test images.

\paragraph{MedThinkVQA~\cite{yao2026medicalthinking}.}
Clinical cases drawn from radiological case reports, each presenting 2 to 47 images with a question
and five candidate diagnoses. The split holds 7,347 training and 720 test cases, and the test cases
carry 5,845 images in total, 8.12 per case on average.

\paragraph{PMC-VQA~\cite{zhang2023pmcvqa}.}
Questions about figures from the biomedical literature, each with four candidate answers written as
free text and not drawn from a label set. We use version 2 of the dataset, whose test split holds
33,430 question-answer pairs over 29,021 figures.

\paragraph{Preprocessing.}
An example is kept when every image of it decodes, when it offers between two and seven candidates, and
when the reference answer matches exactly one candidate. After these filters the four benchmarks
contribute 174,360 training and 36,286 test examples, as listed in Tab.~\ref{tab:config}.

\section{Introduction to the Baselines}
\label{supp:baselines}

We compare against 20 off-the-shelf vision-language models at the 4B to 9B scale, and separately
against six released decision models. Every baseline runs with its official weights and its own chat
template, and each is timed in its own session under the protocol of Sec.~\ref{supp:implementation}.

\paragraph{General Vision-Language Models.}
Fifteen models come from five families.
From LLaVA we take LLaVA-1.5-7B, LLaVA-1.6-Vicuna-7B, LLaVA-1.6-Mistral-7B and
LLaVA-OneVision-7B~\cite{liu2024llava15,liu2024llavanext,li2024llavaonevision}.
From InternVL we take InternVL2.5-4B, InternVL3-9B, InternVL3.5-4B and
InternVL3.5-8B~\cite{chen2024internvl25,zhu2025internvl3,wang2025internvl35}.
From Keye-VL we take Keye-VL-8B-Preview and Keye-VL-1.5-8B~\cite{kwai2025keyevl,yang2025keyevl15},
and from MiMo-VL we take MiMo-VL-7B-RL~\cite{xiaomi2025mimovl}.
From Qwen we take Qwen2.5-VL-7B, Qwen3-VL-4B, Qwen3-VL-8B and Qwen3.5-9B~\cite{bai2025qwen25vl,bai2025qwen3vl,qwen2026qwen35_9b}, four models that span two generations.

\paragraph{Medical Vision-Language Models.}
Four models are trained on biomedical image-text data.
They are LLaVA-Med-v1.5-7B~\cite{li2023llavamed}, HuatuoGPT-Vision-7B~\cite{chen2024huatuogptvision},
MedGemma-4B~\cite{sellergren2025medgemma} and Lingshu-7B~\cite{xu2025lingshu}.

\paragraph{The Backbone.}
Qwen3.5-4B~\cite{qwen2026qwen35} is the model \textsc{MedJev}-4B is built from. It shares the vision
encoder, the connector and the retained language layers, and it is timed in the same session as
\textsc{MedJev}-4B.

\paragraph{Released Decision Models.}
Six systems share the decision interface.
Two of them are medical, OmniMed-Jev-4B~\cite{lytang2026omnimedjev} and
PathJev-P~\cite{tommasino2026pathjev}, both on a MedGemma-1.5-4B backbone.
Four are general-purpose, Visual Jev-4B~\cite{yu2026visualjev} on Qwen3-VL-4B,
d1-3B~\cite{liquid2026d1} on LFM2.5-VL-3B, and Clef-Flash-9B~\cite{cloudflare2026clef} and
d3-flash-9B~\cite{vllmsr2026d3} on Qwen3.5-9B. Each runs
through the entry point its own release documents, at its own resolution and precision, and receives
the candidate texts the benchmark defines.

\paragraph{Answer Extraction.}
A generative baseline returns a string, which is mapped to a candidate by one rule fixed before any
baseline was run and applied unchanged to all of them. A response is accepted when it matches the text
of exactly one candidate, when a line of it reads as an answer line, when its first or last line is a
bare option letter, or when it carries an \texttt{<answer>} tag holding a letter. A response that names
no candidate, names two, or whose letter and text disagree counts as an error.

\section{Further Implementation Details}
\label{supp:implementation}

Tab.~\ref{tab:config} lists the configuration of the deployed \textsc{MedJev}-4B, read from the frozen protocol.

\paragraph{Prompt and Markers.}
A request is laid out as the visual tokens of every image, the question, a question marker, then each
candidate followed by an option marker, and finally a decision marker.
The three markers are added to the vocabulary as \texttt{<|medjev\_\allowbreak question\_\allowbreak end|>},
\texttt{<|medjev\_\allowbreak option\_\allowbreak end|>} and \texttt{<|medjev\_\allowbreak decide|>}, and their embeddings are initialized
from the end-of-turn token of the backbone so that the first forward pass starts from a state the
pretrained model already understands.
Every image keeps all of the tokens the native vision merger produces, up to 1,024 tokens per image and
4,096 per example, and no token reduction or pooling is applied anywhere.

\paragraph{Depth Reduction.}
The backbone of Qwen3.5-4B~\cite{qwen2026qwen35} has 32 language layers.
We keep layers 1 to 23 complete, delete layers 24 to 31 entirely, and keep from layer 32 only the query
and key projections that the matching term needs.
The last layer of Qwen3.5 is a full-attention layer and supplies the matching space.
Its query projection also emits an output gate, which we discard.

\begin{wraptable}{r}{0.46\textwidth}
\vspace{-\intextsep}
\centering
\small
\caption{\textbf{Training and model configuration} of the deployed \textsc{MedJev}-4B, read from
the frozen protocol of the training run that produced the deployed checkpoint of the paper.}
\label{tab:config}
\renewcommand\tabcolsep{3.6pt}
\renewcommand\arraystretch{1.08}
\resizebox{0.44\textwidth}{!}{
\begin{tabular}{l|l}
\Xhline{1.2pt}
\rowcolor{medjevorange!18}
\textbf{Setting} & \textbf{Value} \\
\Xhline{1.2pt}
Retained Language Layers $J$ & 23 \\
Layer Supplying the Matching Projections & 32 \\
Removed Complete Layers & 24--31 \\
Residual Head Hidden Width & 256 \\
Residual Head Activation & GELU \\
Residual Output Initialization & Zero \\
Maximum Candidates per Example & 7 \\
LoRA Rank, Scaling, Dropout & 16, 32, 0 \\
Visual Tokens per Image, per Example & 1,024, 4,096 \\
Maximum Input Tokens & 32,768 \\
Epochs, Global Batch Size, Steps & 1, 32, 5,449 \\
Learning Rate, Weight Decay, Gradient Clip & 0.0001, 0.01, 1 \\
Seed & 42 \\
Training Examples & 174,360 \\
\quad HAM10000 & 9,951 \\
\quad PneumoniaMNIST & 4,698 \\
\quad MedThinkVQA & 7,307 \\
\quad PMC-VQA & 152,404 \\
Test Examples & 36,286 \\
\Xhline{1.2pt}
\end{tabular}}
\vspace{-\intextsep}
\end{wraptable}

\paragraph{Training.}
A single model is trained for one epoch over the 174,360 training examples of the four benchmarks, which
is 5,449 updates at a global batch size of 32.
The loss is a cross-entropy over the candidates of each example, weighted so that no benchmark dominates
and, within each classification benchmark, so that no diagnosis dominates.
Each benchmark first receives the same total weight, which scales an example by the pooled training
size over four times its own, from 4.38 on HAM10000 to 0.29 on PMC-VQA. Within a benchmark, a
classification example is weighted by the inverse of its observed class frequency and
question-answering examples count equally.
The LoRA~\cite{hu2022lora} adapters are merged into the backbone before any timing.

\paragraph{Timing.}
Latency is measured on one H20 at batch size one, with a warmup that is excluded and with the GPU
otherwise idle.
The timed region starts when the request arrives and ends when the decision is on the CPU, so it includes
reading and decoding the image, the processor, the host-to-device copy, the forward pass, and the final
synchronisation.
For the generative baselines the same region also contains the decoding loop and the parser.

\section{Additional Results}
\label{supp:results}

\subsection{Released Decision Models}
\label{supp:released}

\begin{table}[htbp]
\centering
\caption{\textbf{Released decision models, per benchmark.} Accuracy and mean end-to-end latency,
which the main paper summarises as one accuracy column and one latency column. Every row was
measured in one session on one H20 at batch size one, through the entry point the release documents.
The MedThinkVQA columns are scored from the first image of each case, so the input stays the same.}
\label{tab:released_full}
\renewcommand\tabcolsep{4.0pt}
\renewcommand\arraystretch{1.1}
\resizebox{\textwidth}{!}{
\begin{tabular}{l|l|cccc|cccc}
\Xhline{1.2pt}
\rowcolor{medjevorange!18}
& & \multicolumn{4}{c|}{Accuracy (\%)} & \multicolumn{4}{c}{Latency (ms)} \\
\rowcolor{medjevorange!18}
\multirow{-2}{*}{\textbf{Model}} & \multirow{-2}{*}{\textbf{Backbone}}
& HAM & Pneu. & MedThink & PMC & HAM & Pneu. & MedThink & PMC \\
\Xhline{1.2pt}
d1-3B~\cite{liquid2026d1} & LFM2.5-VL-3B & 59.92 & 37.50 & 31.53 & 55.43 & 43.8 & 28.3 & 67.0 & 34.7 \\
Visual Jev-4B~\cite{yu2026visualjev} & Qwen3-VL-4B & 56.81 & 62.02 & 20.56 & 51.99 & 88.5 & 69.5 & 106.8 & 77.5 \\
OmniMed-Jev-4B~\cite{lytang2026omnimedjev} & MedGemma-1.5-4B & 66.67 & 95.03 & 22.22 & 32.93 & 204.0 & 202.8 & 207.1 & 197.4 \\
PathJev-P~\cite{tommasino2026pathjev} & MedGemma-1.5-4B & 51.32 & 64.10 & 26.81 & 37.23 & 194.0 & 195.3 & 196.8 & 186.9 \\
Clef-Flash-9B~\cite{cloudflare2026clef} & Qwen3.5-9B & 61.90 & 79.97 & 28.06 & 56.37 & 112.1 & 61.4 & 124.6 & 75.9 \\
d3-flash-9B~\cite{vllmsr2026d3} & Qwen3.5-9B & 67.00 & 71.47 & 32.92 & 59.88 & 99.8 & 60.3 & 118.5 & 67.7 \\
\textsc{MedJev}-4B (Ours) & Qwen3.5-4B & 68.39 & 94.23 & 51.81 & 61.66 & 64.3 & 50.9 & 260.6 & 55.1 \\
\Xhline{1.2pt}
\end{tabular}
}
\end{table}

Removing generation is necessary but not sufficient, and Tab.~\ref{tab:released_full} is the evidence.
Every model in it already returns scores instead of text, yet none is competitive here, which places
our gain in how the decision is built and not in the interface alone.
The per-benchmark view makes the failure mode explicit. d1-3B lands on exactly 37.50\% on
PneumoniaMNIST, the share of the smaller class, and on 59.92\% on HAM10000, within three cases of the
share of the largest class, so both numbers are what a constant answer would score. A released model
can therefore post a plausible average while deciding almost nothing, and the margin that matters is
the one over the majority-class floor, where \textsc{MedJev}-4B leads by 8.27\% on HAM10000 and
31.73\% on PneumoniaMNIST.

The multi-image column separates the systems further. Every row is scored from the first image of
each case on MedThinkVQA~\cite{yao2026medicalthinking}, which keeps the input identical across
interfaces and leaves 5,125 images unseen, and \textsc{MedJev}-4B still reaches 51.81\% where the best
released model reaches 32.92\%. The deployed model reads every image of a case in one pass, which is
why the design treats the number of images as a property of the request and not as a limit of the interface, and why the
latency is averaged over the three single-image benchmarks and not over all four.

\subsection{Confidence Calibration}
\label{supp:calibration}

\begin{figure}[htbp]
  \centering
  \begin{minipage}{0.49\textwidth}\centering
    \includegraphics[width=\linewidth]{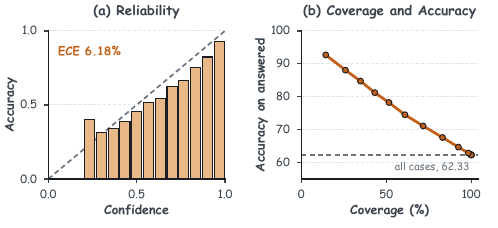}
  \end{minipage}\hfill
  \begin{minipage}{0.49\textwidth}\centering
    \includegraphics[width=\linewidth]{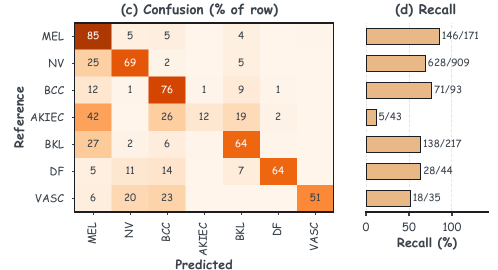}
  \end{minipage}
  \caption{\textbf{Calibration and per-category behaviour of the deployed model.}
  (a) Accuracy against confidence in fifteen equal-width bins over all 36,286 test examples.
  (b) Accuracy on the answered cases when the least confident ones are referred away.
  (c) The confusion on HAM10000, each row normalized by the test cases of that category.
  (d) The recall of each category with its counts.}
  \label{fig:calibration}
  \label{fig:ham}
\end{figure}

Scoring every candidate gives a confidence signal for free, and Fig.~\ref{fig:calibration} (a) shows that the
signal is usable and not decorative. Accuracy rises with confidence across the populated bins
apart from one bin of five examples, and the expected calibration error over all 36,286 test examples
is 6.18\%, so the ordering the head produces indicates whether its decision
should be trusted.

What this buys a deployment is in Fig.~\ref{fig:calibration} (b). Answering only the cases above a
confidence threshold raises accuracy from 62.33\% at full coverage to 92.67\% on the most confident
14.4\% of cases, so the same model can be operated at a chosen reliability by referring the rest to a
clinician. A model that answers in free text gives one answer and no ranking over the
alternatives, so this control costs a pass.

\subsection{Candidate Order}
\label{supp:order}

\begin{table}[htbp]
\centering
\caption{\textbf{Candidate order, in full.} Every test example is decided again with the candidate
order permuted, under three seeds. Kept, Lost and Gained count the natively correct decisions that
survive, those that break, and the natively wrong decisions that become correct, pooled over the
three seeds.}
\label{tab:order_full}
\renewcommand\tabcolsep{3.0pt}
\renewcommand\arraystretch{1.1}
\resizebox{0.86\textwidth}{!}{
\begin{tabular}{l|c|ccc|ccc}
\Xhline{1.2pt}
\rowcolor{medjevorange!18}
\textbf{Benchmark} & $n$ & Native & Permuted (3 Seeds) & Spread & Kept & Lost & Gained \\
\Xhline{1.2pt}
HAM10000~\cite{tschandl2018ham10000} & 1,512 & 68.39 & 68.19 / 67.92 / 67.72 & 0.46 & 3,054 & 48 & 28 \\
PneumoniaMNIST~\cite{yang2023medmnist} & 624 & 94.23 & 94.23 / 94.23 / 94.23 & 0.00 & 1,764 & 0 & 0 \\
MedThinkVQA~\cite{yao2026medicalthinking} & 720 & 51.94 & 49.86 / 50.00 / 49.86 & 0.14 & 975 & 147 & 103 \\
PMC-VQA~\cite{zhang2023pmcvqa} & 33,430 & 61.68 & 60.19 / 60.48 / 60.25 & 0.28 & 56,015 & 5,845 & 4,464 \\
\Xhline{1.2pt}
\end{tabular}}
\end{table}

\begin{figure}[htbp]
  \centering
  \includegraphics[width=\textwidth]{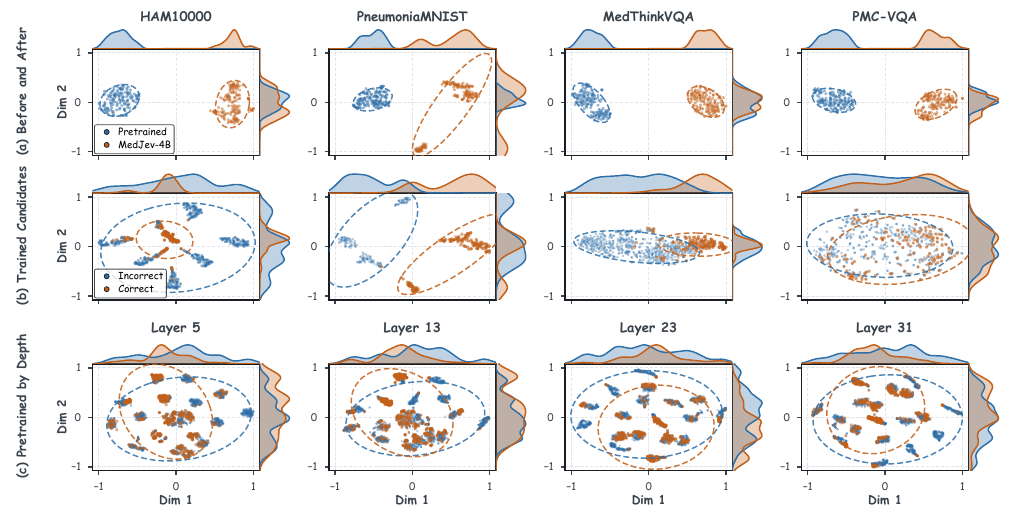}
  \caption{\textbf{Hidden representations in detail.} Each panel is reduced on its own, so its layout
  is never driven by the others. (a) and (b) give one column per benchmark, the decision marker state at
  layer 23 before and after training, and the candidate marker states of the trained model split by
  whether the candidate is correct. (c) gives the same split at four depths of the pretrained backbone.}
  \label{fig:repr_detail}
\end{figure}

\begin{table}[htbp]
\centering
\caption{\textbf{Per-request latency and generated tokens.} Median and 99th percentile of the
end-to-end latency in milliseconds, and the median number of generated tokens, over 150 timed requests
per benchmark and model on one H20 at batch size one. Warmups are excluded, thinking is off, and the
three requests MedGemma-4B failed on PMC-VQA are left out of its own statistics.}
\label{tab:latency_full}
\renewcommand\tabcolsep{4.2pt}
\renewcommand\arraystretch{1.1}
\resizebox{\textwidth}{!}{
\begin{tabular}{l|ccc|ccc|ccc|ccc}
\Xhline{1.2pt}
\rowcolor{medjevorange!18}
& \multicolumn{3}{c|}{HAM10000} & \multicolumn{3}{c|}{PneumoniaMNIST}
& \multicolumn{3}{c|}{MedThinkVQA} & \multicolumn{3}{c}{PMC-VQA} \\
\rowcolor{medjevorange!18}
\multirow{-2}{*}{\textbf{Model}}
& Med. & p99 & Tok. & Med. & p99 & Tok. & Med. & p99 & Tok. & Med. & p99 & Tok. \\
\Xhline{1.2pt}
MedGemma-4B~\cite{sellergren2025medgemma} & 214 & 223 & 3 & 206 & 213 & 3 & 809 & 3317 & 3 & 213 & 223 & 3 \\
Qwen3.5-4B~\cite{qwen2026qwen35} & 208 & 457 & 4 & 260 & 311 & 6 & 354 & 27814 & 4 & 361 & 790 & 8 \\
\textsc{MedJev}-4B (Ours) & 72 & 84 & 0 & 62 & 78 & 0 & 207 & 625 & 0 & 63 & 81 & 0 \\
\Xhline{1.2pt}
\end{tabular}
}
\end{table}

A model that scores candidates could be reading the slot a candidate occupies instead of the text in
it, and Tab.~\ref{tab:order_full} bounds how much of that is happening. Decided again under three
permuted orders, the unweighted mean falls from 69.06\% to 68.10\%, the largest single drop is
2.03\% on MedThinkVQA, and PneumoniaMNIST does not move at all, so position accounts for at most a
small part of it.

The churn underneath that stability is the more informative part. On PMC-VQA the permutation costs
5,845 correct decisions and returns 4,464 that were wrong before, a net loss of 1,381 out of 100,290
draws. The two directions are close in size, so the order rearranges which examples are decided correctly far more than it changes how many of them there are.

\subsection{Per-Category Behaviour on HAM10000}
\label{supp:ham}

An average over seven categories can hide where a model fails, so Fig.~\ref{fig:ham} (c) and (d) open it up, and
the accuracy turns out to be concentrated where the clinical cost is highest. \textsc{MedJev}-4B
recalls melanoma in 146 of 171 cases and basal cell carcinoma in 71 of 93, while still recalling the
dominant benign category of melanocytic nevus in 628 cases out of 909.

The errors also fall in the direction a screening setting prefers. The confusion is asymmetric towards
the malignant categories, with 25\% of the nevus cases sent to melanoma and far fewer in the opposite
direction, so the errors that remain fall mostly on the side of over-referral. That asymmetry belongs
to the decision and not to the interface, since every category is in the candidate set and always scored.

\subsection{Per-Request Latency}
\label{supp:latency}

A mean latency hides the behaviour a deployment has to budget for, and
Tab.~\ref{tab:latency_full} shows that the two interfaces differ in kind. For the
generative models the 99th percentile runs far above the median, because the cost is set by the response length, reaching tens of seconds on MedThinkVQA.

\textsc{MedJev}-4B has almost no tail, and its median and its 99th percentile sit within twenty
milliseconds of each other on the three single-image benchmarks, while the multi-image MedThinkVQA
cases stay inside one second at the 99th percentile. The work it does is fixed by the request and not
by the answer it is about to give, which makes the cost of a decision predictable in advance.

\section{Additional Visualisations}
\label{supp:visual}

The main paper reduces all four benchmarks together, which is the right view for its claim that
training moves the four sources into one shared region.
Fig.~\ref{fig:repr_detail} (a) and Fig.~\ref{fig:repr_detail} (b) reduce each benchmark on its own
with t-SNE~\cite{maaten2008tsne} instead, so that a panel cannot be arranged by the other three.
The shift from the pretrained states to the trained ones holds inside every benchmark, and the split
between correct and incorrect candidates is visible in each of them, which rules out the reading that
the separation in the pooled figure is an artefact of the benchmarks sitting in different regions.

Fig.~\ref{fig:repr_detail} (c) looks at the pretrained backbone alone, with none of our training
applied.
The correct candidates are already pulled away from the incorrect ones at depths well below the top of
the stack, and the geometry does not sharpen much over the last layers.
This is the same observation the depth probe of the main paper makes with a linear readout, seen here
without fitting anything, and it is the reason why a model that keeps only 23 layers loses no accuracy.

\end{document}